# Blockchain for Multi-Robot Collaboration to Combat COVID-19 and Future Pandemics

## S. H. Alsamhi, Brian Lee

[1] IBB University, Ibb, Yemen

Correspondence: (s.alsamhi.rs.ece@iitbhu.ac.in)

**ABSTRACT** This conceptual paper overviews how blockchain technology is involving the operation of multi-robot collaboration for combating COVID-19 and future pandemics. Robots are a promising technology for providing many tasks such as spraying, disinfection, cleaning, treating, detecting high body temperature/mask absence, and delivering goods and medical supplies experiencing an epidemic COVID-19. For combating COVID-19, many heterogeneous and homogenous robots are required to perform different tasks for supporting different purposes in the quarantine area. Controlling and decentralizing multi-robot play a vital role in combating COVID-19 by reducing human interaction, monitoring, delivering goods. Blockchain technology can manage multi-robot collaboration in a decentralized fashion, improve the interaction among them to exchange information, share representation, share goals, and trust. We highlight the challenges and provide the tactical solutions enabled by integrating blockchain and multi-robot collaboration to combat COVID-19 pandemic. The framework of our conceptual proposed can increase the intelligence, decentralization, and autonomous operations of connected multi-robot collaboration in the blockchain network. We overview blockchain potential benefits to defining a framework of multi-robot collaboration applications to combat COVID-19 epidemics such as monitoring and outdoor and hospital End to End (E2E) delivery systems. Furthermore, we discuss the challenges and opportunities of integrated blockchain, multi-robot collaboration, and the Internet of Things (IoT) for combating COVID-19 and future pandemics.

## I. INTRODUCTION

Recently, the World Health Organization (WHO) is considered coronavirus disease 2019 (COVID-19) as an international pandemic epidemic. In July 2020, WHO reported that the number of confirmed COVID-19 cases reached 14 million people with 540,000 dead rates [1]. COVID-19 has the capability to spread very quickly. The most efforts to control the spreading COVID-19 have failed; global monitoring of infected people is highly demanded. Lock-down can reduce the number of infected people of COVID-19 due to separating the infected people from others (avoid interaction). Furthermore, diagnosis and early detection may lead to less infected people and better healthcare to infected patients [2]. Advanced technologies have shown their capabilities in combating COVID-19 outbreak in early diagnosis, monitoring infected patients, and minimizing the interaction between staff and infected patients in the hospitals [3]. Several robots have been used for combating COVID-19 outbreak (autonomous robots, telemedicine robots, and collaborative robots, mobile robots) to avoid interaction between humans, monitoring

(social distancing, discover new COVID-19 cases), delivering goods, spray disinfection, etc.[4]. So, a massive number of robots are required for performing different tasks to combat COVID-19 in the quarantine area and hospital simultaneously.

Network management of multi-robot is tedious due to the requirement of homogenous and heterogeneous multi-robot to collaborate effectively efficiently. Managing multi-robot collaboration requires for improving energy efficiency [5], enhancing Quality of services (QoS) [6], avoiding collision[5, 7], delivering services efficiently [8] during combating COVID-19 outbreak. For instance, homogenous multi-robot collaboration performs the same tasks (i.e., medical/food delivery robots, cleaning robots in hospitals), while heterogeneous robots are required to perform several tasks in different environments. Robots can collaborate with smart devices to perform tasks efficiently, such as collaboration of robots and the Internet of Things (IoT) in smart cities applications [8]. However, managing a multi-robot network needs protocols to perform their complex tasks efficiently. Furthermore, multi-robot needs to collaborate in sharing location, purpose, and tasks.

With the support of IoT, Cloud/Edge computing, and Artificial Intelligence (AI), multi-robot has become adaptive, collaborative, and connected, thereby enabling complex interactions amongst robots. Multi-robot collaboration interacts with each other to exchange information, share representation, goals, independent subtasks, and mutual learning adaption and trust. Multi-robot collaboration may face the disconnectivity issue because of communication challenges or navigational during task performance. For instance, malfunctioning in centralized multi-robot collaboration stops updating all robots to collaborate based on sharing information and goals. Managing tolerance network partitioning requires for decentralized robot collaboration. So, if robot malfunctioning occurs, other robots in the swarm will continue sharing and update their actions in order to perform tasks efficiently. However, many challenges still exist to impede the widespread adoption of multiple large-scale robots (e.g., swarm robotics [9] including robots network architecture, supervision of robots, scalability, controlling, network partitioning, malfunctioning, and time and energy efficiency. Many techniques are used to solve existing challenges such as learning, distributed decision making, blockchain, etc.

Distributed decision-making algorithms (DDM) have been applied in the development of robotics applications such as collection mapping [10], dynamic task allocation [11], avoiding obstacle [12]. However, DDM algorithms are an open problem in deploying massive numbers of robots [13]. Furthermore, the flexible and autonomous multi-robot decision making by using DDM algorithms are needed to tackle the new wave of challenges facing the industry. However, blockchain is ensuring that all robots in a decentralized network are sharing an identical state. For example, blockchain creates a distributed voting among multi-robot, that is required to reach an agreement.

Moreover, blockchain technology achieves collaboration models between heterogenous robots in a multi-robot system [14]. Further, blockchain technology adoption in distributed decision making can provide operators and maintainers to multi-robot collaboration. Furthermore, blockchain is used widely over AI due to unique advantages in surveillance [15]. Blockchain advantages are including decentralized, immutable, deterministic, data integrity [16]. Because of all transactions and agreements store in blockchain, there is no requirement for investing time in training and learning phases in joining new robots to multi-robot collaboration [14]. Then, new robots can synchronize automatically with multi-robot via downloading the historical events of all previously stored from the blockchain.

Various works have proposed using blockchain to facilitate robot collaboration [14, 17, 18]. With blockchain, independent robots can reach consensus without a central controller. Smart contracts provide great potential to enable more secure [17], autonomous, flexible, and even profitable [19] robotic operations [20]. Combating COVID-19 outbreak requires automatic robots to collaborate with each other in a decentralized fashion without human intervention during task performance. Task performance needs a dynamic data collection from the quarantine area. Therefore, the collaboration of multi-robot automation is required for decentralized networks to increase the efficiency to combat COVID-19. Here, blockchain can store data efficiently, enhance security, and share automation transactions between autonomous multi-robot collaboration [21]. Furthermore, blockchain helps robots to participate in performing actions, making decisions, planning activities collaboratively. Therefore, the combination of blockchain and automated multi-robot collaboration is essentially required for combating COVID-19.

Figure.1 illustrates the advantages of blockchain technology for solving multi-robot collaboration issues to combat COVID-19 outbreak. There are several ways for robots to avoid COVID-19 outbreak by monitoring and guiding people in the quarantine area, avoiding interaction in the hospital, and avoiding interaction in the quarantine area. In case of monitoring and guiding people, robots are used to produce an alarm to sperate people gathering, detect people with the absent mask, spray disinfection, detect high temperatures of the infected person of COVID-19 disease. To avoid interaction, robots are used in hospitals to spray disinfection, cleaning, testing, and End to End (E2E) system for delivering food/medical. While avoiding interaction in the quarantine area, robots are used for End to End (E2E) delivery (i.e., medicine, foods, and sample tests, etc.), disinfection. In the above cases, blockchain is used to turn alert, share information, decentralization, update robot and load status, share location during robot navigation, etc. Consensus algorithms and sharding techniques are required to improve the performance of blockchain in the transaction speed, scalability, network partitioning, and malfunctioning robots. Therefore, multi-robot can collaborate to combat COVID-19 effectively and efficiently indoor (hospital) and outdoor (quarantine area).

The current research is still at an early stage. The focus is mainly to verify how to apply blockchain technology in managing multi-robot that can perform different tasks for different purposes in combating COVID-19 outbreak. While the results so far are positive, many open challenges address here. Most solutions require each robot to process transactions and store the world state equally, just as what the normal blockchain systems do. This makes the research on blockchain for improving multi-robot collaboration as a very relevant and strategical topic for combating COVID-19 outbreak.

## A. MOTIVATION
Practicing social distance, avoiding interaction, and staying at home are ways to reduce infected people of COVID-19 and spread COVID-19 outbreak. To apply them in real-world (in a quarantine area and hospital), quarantined people needs for emerging technologies to participating in (i) E2E delivery system, (ii) monitoring gathering people and detecting absence mask (iii) detecting infected people from long distance, (iv) spray disinfection, etc. Robots are used for different purposes (spraying, disinfection, monitoring, cleaning, detecting, medical test, delivery goods) in order to combat COVID-19 outbreak. In the hospital, there are several robots used, such as cleaning robots, testing robots, spray disinfection robots, delivering medicine, and food robots. To perform tasks collaboratively and efficiently in real-time, multi-robot needs to be managed and controlled to success their tasks without collision or delay.

Furthermore, a multi-robot collaboration network, centralized control suffers from a single point of failure, whereas decentralized control suffers from lacking global knowledge. Therefore, decision making in the centralization robots network takes a long time for controlling multi-robot during task performance, and the collision between robots may occur due to delay in the time response. Decentralized improves the performance of the robots and reduces the time spent on doing tasks. Furthermore, sharing information is essential to support multi-robot collaboration for operation in environmental exchange, uncertain conditions, and external disturbances. A successful solution for multi-robot interaction issues to perform tasks and record event history by blockchain can improve the efficiency of multi-robot interaction during combating COVID-19 outbreak in the hospital and the quarantine area. Therefore, implementing blockchain in multi-robot is required to increase interaction between robots via supporting information exchange with high trust, detecting malfunctioning, helping robot detecting performance issues, reaching a consensus, deploying distributed solutions, allocating plans and tasks, and joint missions.

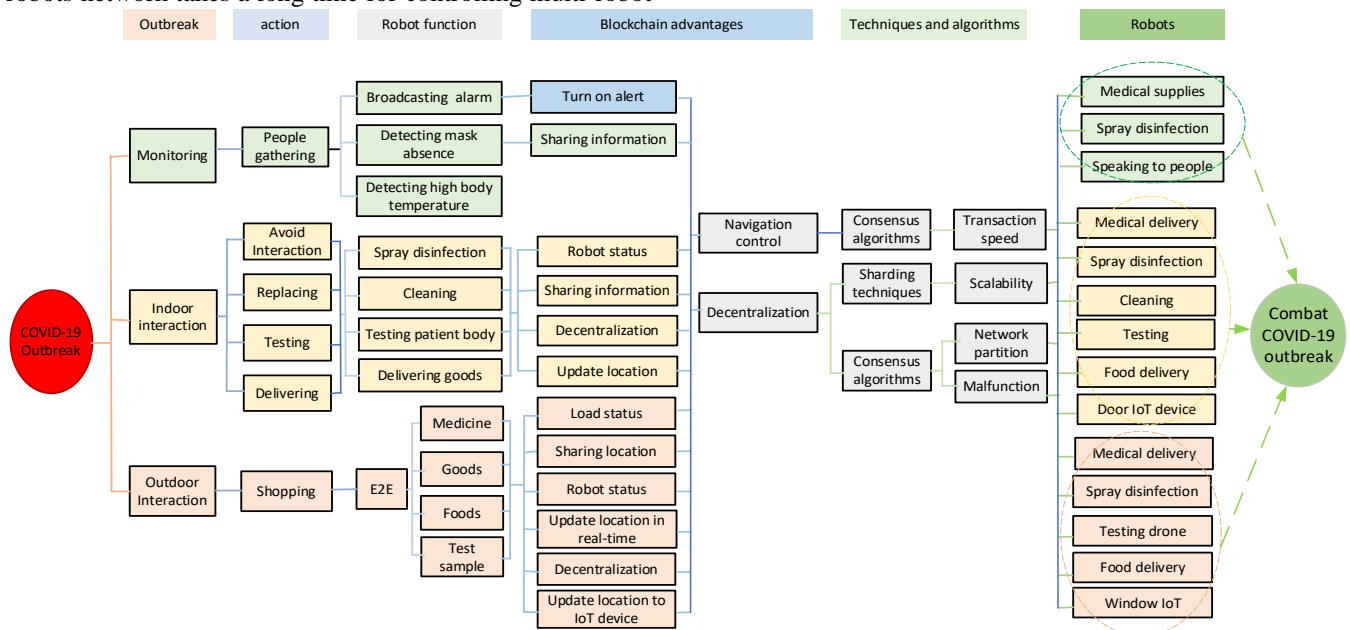

Figure.1 Blockchain solutions for improving multi-robot collaboration to combat COVID-19

## B. CONTRIBUTION AND SCOPE

CoVID-19 outbreak is rapidly and becomes very dangerous due to killing so many people in the world. The convergence of several advanced digital technologies plays a critical role in combating and controlling the outbreak of COVID-19. Therefore, using robots for combating COVID-19 makes the research on blockchain for improving multi-robot collaboration as a very relevant and strategical topic. For combating COVID-19 outbreak efficiently and effectively, managing homogenous and heterogenous robots, collaboration is necessary to avoid a collision, reduce response time, speed transaction, aid scientists to detect and identify infected cases earlier. Furthermore, due to increasing the number of robots for combating COVID-19 in different ways in need, robots are suffering from global information, collision, malfunction, controlling collaboration, and so on.

This conceptual research proposes to focus on the avoidance network partitioning, improving the scalability of multi-robot, controlling the performance, which can contribute in a clear and more effective in combating COVID-19 tasks in the quarantine area. As the difference of previous researches on the same topic, this research mainly focuses on blockchain for managing a multi-robot collaboration in order to combat COVID -19 outbreak in the quarantine area. Blockchain makes data public for all robots in a network, enables robots to log information, and delivers robot relation to all neighbors' robots. We focus mainly on discussing challenges in multi-robot collaboration and how to develop a decentralized ledger platform with corresponding algorithms to enable multi-robot collaboration with a tolerance of network partition and malfunctioning robots as follow:

- **Multi-robot collaboration management and controlling:** We discuss the proposed blockchain

technology as a critical solution framework for managing and controlling multi-robot collaboration to combat COVID-19 outbreak. Blockchain is proposed to control homogenous and heterogeneous robots to avoid collision during operation in uncertain conditions and environment exchanges. Each robot is sharing information with others via a smart contract. Then, robot action and behavior can be changed accordingly, until performing a specific task (combating COVID-19) effectively and efficiently. Robotchain is decentralized, which can handle a large number of transactions that are produced from multi-robot per second. Therefore, it notes that the interaction among multi-robot collaboration can increase efficiency to combat COVID-19.

- **Decentralized of multi-robot collaboration:** Blockchain is a decentralization network in which multi-robot collaboration suffers. Moreover, blockchain helps the robots to process the transaction and store the world state equally. In the case of a malfunctioning robot, multi-robot collaboration can perform their task efficiently. Furthermore, joining a new robot is more accessible by copying the smart contract and starting sharing events in the blockchain. Due to the decentralization feature which offers by blockchain, both joining a new robot and malfunctioning robot cannot affect the performance of multi-robot collaboration to combat COVID-19. Here, we discuss the characteristics of realistic multi-robot collaboration for combating COVID-19 and how to develop the consensus algorithms with considering consistency based on the current status. Further, we discuss the consensus algorithms with dynamic sharding technique to increase blockchain scalability, so that the transaction processing and world state is limited in the sharing range.

- **Monitoring quarantine area:** We explore the proposed framework of multi-robot collaboration application for detecting high body temperature, spraying disinfection, and detecting mask absence during monitoring quarantine area. In this framework scenario, blockchain is used for sharing information to support the interaction of heterogeneous multi-robot collaboration. For detecting high body temperature and mask absence, detection protocol can be prepared by deep learning and stored in a smart contract. In the case of mask absence detection, the robot will update location and register in the blockchain, while the blockchain turn on alert of guiding the person to follow the restrictions of quarantine. In the case of high body temperature detection, the robot location will be updated to the blockchain. The spray infection robot will come and perform spraying disinfection, announce robot will speak the guideline, while the self-driver robot will take an infected case to the hospital for ensuring infected cases by using a testing robot.

- **Quarantine E2E delivery system**: We address the proposed framework solution for delivering medical supplies and foods/goods to the quarantine area using blockchain and multi-robot collaboration. Furthermore, the proposed framework architecture consists of the donate layer (market, pharmacy), blockchain layer, and received layer/customer layer (order goods and medicine). The transport robot carries order to customers based on blockchain encrypt data (path, receiver, robot ID, and location). A smart contract verifies and executes robot status between denoting and customer/receiver. The robot performs the transport task and keeps updating location to the smart contract in blockchain for notifying denote and customer/receiver. As soon as the transport robot delivers an order, it will update the status "delivery reaches successfully" to the blockchain.

- **Hospital E2E delivery system:** We address blockchain and multi-robot collaboration as proposed framework solutions for the E2E delivery system inside the hospital. The proposed framework solution can serve quarantined people in the hospital without hospital staff interaction. To avoid collision and decrease interaction between hospital staff and patients, all robots must be connected to the blockchain network. A medicine delivery or food delivery robot updates its location to the blockchain, while a smart contract shares it with all robots in the blockchain network until the robot reaches the goal. In smart hospitals, smart IoT devices at the door must be the node of the blockchain network and get updates of the specific robot from a smart contract. Smart IoT devices can help if a patient can not open the room door due to the critical case of COVID-19. It notes that the combining of blockchain, multi-robot collaboration, and smart IoT technologies can satisfy the patient quality of experience and combat COVID-19 by reducing human interaction in the hospital.

The rest of the paper is organized, as shown in Figure.2. Multi-robot collaboration and COVID-19 and blockchain for multi-robot collaboration are discussed deeply in sections II and III. Section IV and V are introduced to the proposed framework solutions and application domains, while discussion and conclusion are presented in sections VI and VII, respectively.

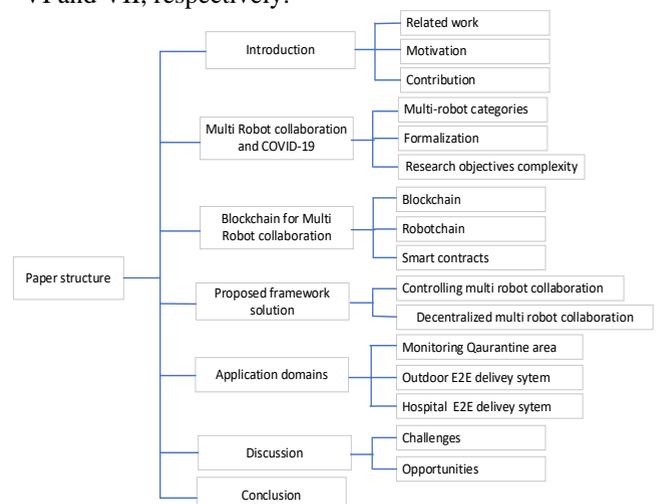

Figure.2 Paper structure

## II. RELATED WORK

Digital technologies play a vital role in fighting COVID in the early beginning, monitoring during COVID-19 outbreak and recovery. Advanced digital technologies include IoT, robots, AI, big data, drone, 5G, cloud computing, and blockchain. The above-mentioned technologies are used for surveillance, monitoring, prevention, and detection of COVID-19 [4, 22-26]. IoT technology is considered for combating the COVID-19 outbreak [27]. The authors of [28] applied blockchain for multi-robot to reach consensus even if Byzantine robots exist. Moreover, the authors of [29] discussed how the combination of blockchain and IoT are helping infected people of COVID-19 disease. IoT devices are allocated in the infected person body for remote monitoring, while blockchain is used to secure gathered information and regulate the medical distribution network. The authors of [30] introduced robotics and smart wearable to facilitate containment, prevention, and mitigation COVID-19. The combination of robotics, smart wearable, AI, and autonomy can meet the potential system, safer and more efficient for care delivery. Furthermore, the authors of [31] described robotics role in the healthcare domain to control the spreading COVID-19. Also, the authors of [32] addressed the decision making for public safety management, which is suitable for controlling COVID-19 outbreak.

The communication among multi-robot is required for data collection, while AI is considered for processing data efficiently [33]. The authors of [34] introduced game theory for wireless communication based on gathered data from different environments. For gathering data from the robot environment, adaptive blockchain technology is required for power consumption and used for improving the battery life of robots during fighting COVID-19 outbreak [35]. The authors of [36] introduced a multi-group SEIRA model for spreading COVID-19 among heterogeneous features of populations. In [37], studied the dynamics spreading of COVID-19 in Brazil (Rio de Ja Neiro state) using a susceptible-infectious-quarantine-recovered model. The results showed that social distancing rules led to a decrease in the rate of growing COVID-19 cases. For modeling COVID-19, the authors of [38] applied a non-singular fractional-order derivative to investigate the model for calculating the transmissibility of COVID-19 disease. In [39], the authors discussed the forecasting and modeling of epidemic COVID-19 and beyond. For tracking COVID-19, blockchain plays a vital role in ensuring the transparency of the epidemic information, the epidemic materials traceability [40]. The authors of [41] highlighted the challenges which have been raised during COVID-19 outbreak. Furthermore, they addressed blockchain as a key technology for solving highlighted challenges (i.e., contact tracing, patient information sharing, disaster relief, automated surveillance, etc.).

Blockchain adapted to support interacting and changing the behavior of multi-robot in the network to perform many events quickly [42]. The authors of [43] proposed blockchain-based decentralized and secure connected autonomous vehicles architecture. For robot supervision and control, it is difficult to collect data from a large number of robots. Consequently, robot supervision using partial knowledge extracted from the data may be error-prone. However, blockchain uses to improve the global data of the multi-robot system, which leads to easier maintenance and higher productivity. Multi-robot systems have wide application scenarios, especially in dangerous, unknown, unreached, or hazardous environments, e.g., humanitarian demining, search and rescue, public safety management [32], surveillance [17], and quarantine area [44]. The consensus problem is fundamental in multi-robot coordination to achieve a common view on the world or to converge to a single decision [45]. In many scenarios, centralized control is not practical, and robots have to communicate with each other in a peer-to-peer manner. Managing multi-robot collaboration for different purposes by using blockchain is essential to perform specific tasks effectively and efficiently in real-time [46].

In [14], the authors presented the benefits of integrating blockchain into a multi-robot system, including security, distributed decision making, robots, behavior differentiation, and new business models. Furthermore, blockchain technology makes robots able to work in dynamic environments without change to the behavior, and control robots for legal responsibility, and ensure safety in each robot in aswarm robots. The authors of [18] presented a blockchain framework as a communication scheme for securing a ride-sharing service between passenger and autonomous vehicles trustworthy and dependable. In [17], the authors proposed an Ethereum based collective decision-making approach for swarm robotics systems (with byzantine robots) in searching for a best-of-2 problem. The approach is compared with a classical approach, i.e., Probabilistic Finite State Machine (PFSM) [47]. The results showed that the classic algorithm breaks down even when the number of byzantine robots is small, whereas the blockchain-based approach provides a robust solution. Strobel [17] introduced a Proof-of-Concept (PoC) that improve the security and stability of the swarm robots coordination and identify byzantine robots from the swarm by using a smart contract of blockchain technology. Also, the decision making of the robots collaboration performance was discussed in the absent and present of byzantine robots.

Blockchain can provide a transparent and immutable robot network that can efficiently mitigate attacks and threats. Blockchain plays a vital role in multi-robot applications to improve interaction efficiency between robots via supporting information exchange with high trusted, reach consensus, change robot behavior and tasks, and deploy joint missions and distributed solutions [48]. Blockchain applications, challenges, and opportunities are discussed with details in [49-51]. A blockchain can also serve as a shared knowledge base and audit log [52] that can be analyzed to improve the system performance. Furthermore,

Table.1 Multi-robot collaboration and COVID-19 challenges with blockchain efforted solutions

| Ref. | Issues | Highlighted | Blockchain | Robot | Multi-robot | IoT | COVID-19 | Importance of blockchain |
|---|---|---|---|---|---|---|---|---|
| [4] (2020) | COVID-19 outbreak | Managing the impacts of COVID-19 and the role of advanced technologies | √ | √ | | √ | √ | Share information |
| [23] (2020) | COVID-19 outbreak | AI and blockchain to combat COVID-19 | √ | | | | √ | Offer solutions for outbreak tracking, safe operations, user privacy protection, etc. |
| [56] (2020) | Heterogenous robots | Blockchain for managing heterogeneous multi-robot collaboration | √ | √ | √ | | | • Share heterogeneous data robots with different computing resources and sensing capabilities |
| [22] (2020) | COVID-19 | Digital technologies for surveillance, monitoring, prevention, and detection COVID-19 | √ | | | √ | √ | • Deliver infected person medication to his doorsteps<br>• Ensure the time of medication delivery along with tracking accuracy. |
| [27] (2020) | COVID-19 | IoT technologies in combating COVID-19 | | √ | | √ | | |
| [29] (2020) | COVID-19 | IoT and blockchain for combating COVID-19 disease | √ | | | √ | √ | • Secure transfer infected person data<br>• Regulate medical distribution network |
| [61] (2019) | Transaction speed and control | Tezos technology is used for enhancing security, while AI in a smart contract to improve robots quality and performance | √ | √ | | | | • Improved security by Tezos<br>• Prevent unwanted behavior changes in robot actions<br>• Improving robot automation |
| [55] (2019) | Data storage and resources computing | Multi-robot and decentralized blockchain to improve Cyber-Physical systems | √ | √ | √ | | | • Decentralization, scalability, consensus, trust, immutability, transparency, autonomous, accessible, and consistent. |
| [57] (2020) | Low-speed transaction | Consensus algorithms improve transaction speed such as DPBFT | √ | √ | | | | • DPBFT consensus algorithm improves real-time transactions in the blockchain |
| [17] (2018) | Security Control | Blockchain is managing security in multi-robot, while the developing of decentralized technique is based on smart contracts | √ | √ | √ | | | • Secure multi-robot coordination with identifying and eliminate the Byzantine robot<br>• Managing Byzantine robot<br>• Collective decision-making |
| [52] (2018) | Malfunction Control | Consensus algorithms in multi-robot are to achieve a fully decentralized fashion. | √ | | √ | | | • Shared knowledge, computing, and reputation management in the multi-robot system |
| [58] (2019) | Network partitions | SwarmDAG for managing a distributed ledger | √ | | √ | | | • Tolerating network partitions<br>• Distributed ledger maintenance |
| [41] (2020) | COVID-19 | Blockchain applicability as enabling technology for solving the challenges that have arisen during COVID-19 pandemic | √ | √ | | √ | √ | • Support contact tracing, patient information sharing, disaster relief, automated surveillance, manufacturing management, contact-less delivery etc. |
| [14] (2018) | Collision Latency Transaction speed | Combination of blockchain and multi-robot for data confidentiality and entity validation; efficiently designed and implemented, and change robots behavior | √ | | √ | | | • Robots reach consensus on a particular state of affairs<br>• Make robots operations autonomous, more secure, profitable, and flexible.<br>• Increase robots flexibility and decreasing complexity |
| [30] (2020) | COVID-19 | Robotics and smart wearable to facilitate the Containment, prevention, and mitigation COVID-19 outbreak | | √ | √ | | √ | |
| [31] (2020) | COVID-19 | Managing COVID-19 by exploring healthcare innovation based on robotics utilization | | √ | √ | | √ | |
| Work | COVID-19 Delivery goods Collision Centralization Supervision Scalability Navigation Detection | Blockchain for managing multi-robot collaboration to combat COVID-19, while consensus and sharding are designed in a smart contract to develop and improve blockchain technology. | √ | √ | √ | √ | √ | • Avoid collision by managing multi-robot<br>• Detecting high body temperature or absence mask<br>• E2E delivery system indoor and outdoor<br>• Improving blockchain decentralized feature-based smart contracts, consensus algorithms, and sharding techniques. |

In [53], it focused on access control of decentralized robotic networks to create a trusted management model with enhancing reputation for each node. RobotChain [54] a decentralized ledger for recording robot events; the smart contract was managing multi-robot with processing data by Oracle. The advantages of blockchain for managing a multi-robot system should fulfill decentralization, scalability, consensus, trust, immutability, transparency, autonomous, accessible, and consistent [55].

The authors of [56] explored the need for blockchain technology for managing the collaboration in heterogenous multi-robots. Furthermore, consensus algorithms were discussed with disadvantages for each algorithm, while sharding techniques were explained to improve the scalability of the blockchain network. The authors of [57] introduced consensus algorithms in order to improve transaction speed applications. They found that the dynamic-reputation Practical Byzantine Fault Tolerance (DPBFT) algorithm was efficient for real-time transactions of consensus algorithms in the energy domain. In [9], the authors implement a distributed ledger system for swarm robotics using BigchainDB to demonstrate the feasibility. The system includes Parrot AR Drone 2.0 and Parrot Jumping Night ground drones. In [58], the authors propose a partition tolerant distributed ledger protocol for swarm robotics named SwarmDAG. SwarmDAG uses a Directed Acyclic Graph (DAG) structured distributed ledger to enable robot swarms to achieve eventual consistency. It employs memberships in different partitions to achieve consensus within each partition. Furthermore, BigchainDB [59] provides higher transaction speed to solve scalability than Bitcoin blockchain [60]. However, machine learning (ML) processes transactions for a long time to run, and therefore, transaction speed by using blockchain becomes negligible. Table.1 shows multi-robot collaboration and COVID-19 issues and blockchain efforts the best solutions. No work has been done based on blockchain for multi-robot collaboration to combat COVID-19 outbreak.

Robotchain [61] proposes a robot action recording system by using a private Tezos blockchain. The recorded events can be deciding robot actions in conflicting situations, monitoring, and tuning robot performance. DezCom [62] proposes a decentralized message broker for industry 4.0 scenarios focusing on robot collaborations. The system is implemented on top of robots running the Robot Operating System (ROS) using the Tendermint consensus protocol. In [63], the authors described the mapping protocol of blockchain for distributed multiple robot systems. The protocol developed based on designing an embedded system with considering options and trade-off for implementing blockchain. Furthermore, blockchain was integrated with swarm robotics communication to avoid network partitions due to communication channels or robot navigation in an experiential environment to perform tasks efficiently [58]. SwarmDAG enabled to the management of network partitions and maintained a distributed ledger. The benefits of using blockchain in swarm robotics were discussed in [17], which are byzantine agents and data robustness structure to wrong data. In [61] introduces blockchain technology as a suitable solution to prevent unwanted behavior changes in robot actions. The blockchain ledger improved the productivity of the factory. The experimental work was in the platform of Tezos blockchain for robot automation and capability. Blockchain is an efficient securing transaction, which allows the decentralization of information in homogenous and heterogenous registries among all homogenous and heterogeneous robots.

The existing works need homogenous and heterogenous robots with different processing and storage capabilities in the blockchain network to perform different tasks to combat COVID-19. However, multi-robot requires the collaboration concept to void the challenges and perform the tasks in real-time effectively and efficiently. The challenges of multi-robot collaboration include global knowledge, network partitioning, scalability, and controlling. This conceptual framework addresses multi-robot collaboration by using blockchain.

## III. MULTI-ROBOT COLLABORATION AND COVID-19

Robots could reduce the cases of infection, and mortality among health care staff, hoping to provide efficient medical assistance. Robots are being deployed to combat COVID-19 via enforcing quarantine restrictions by delivering medical and food, cut the disease risk of interaction between people, and remotely disinfect hospitals [64]. In China, robots are deployed in private and public places, including Guangzhou, Shanghai, and Guiyang, to detect absent masks and high body temperature [65]. For example, a robot carries a loudspeaker to guide individuals remotely in the quarantine area [66]; a high-resolution camera to detect absence mask [67]; a thermal camera to detect body temperature and identifying infected of COVID-19 [68, 69]; and delivering goods and medical supplies (i.e., foods, medicine, and transport patient samples from hospital to disease center, transport medical supplies, and test samples) [70]. Previously, it has been used to prevent diseases outbreak such as Malaria (detect, catch, and take blood samples of mosquitos) [71].

Collaboration is a new concept in which every robot in group robots share its idea about the environment and event, as shown in figure.3. It illustrates the collaboration between robots based on exchange information, sharing representation, share the task, independent sub-tasks, and adaptation, and trust. Interaction among robots relies on information exchanged. Each robot operates in separate sub-tasks independently. The action performance is independent of each robot agreement. Collaboration occurs when every robot depends on other robots to succeed and reach the goal effectively and efficiently. Autonomous multi-robot collaboration can perform tasks, build plans without interventions of humans, which represent can apply with unlimited applications. Therefore, autonomous multi-robot collaboration is required to combat COVID-19 due to their operations without humans intervention. The advantages of autonomous multi-robot collaboration include easily scalable, failure of robot would not result in the group and performing task, better adaptability to the environment, less memory, less processing power, better options for exception controlling, and less capability for individual robots. The authors of [72] applied a collaborative control approach for robot collaboration to perform a common task and achieve a common goal without centralization control. Furthermore, activities and tasks for more than one robot are recognized effectively and efficiently [73].

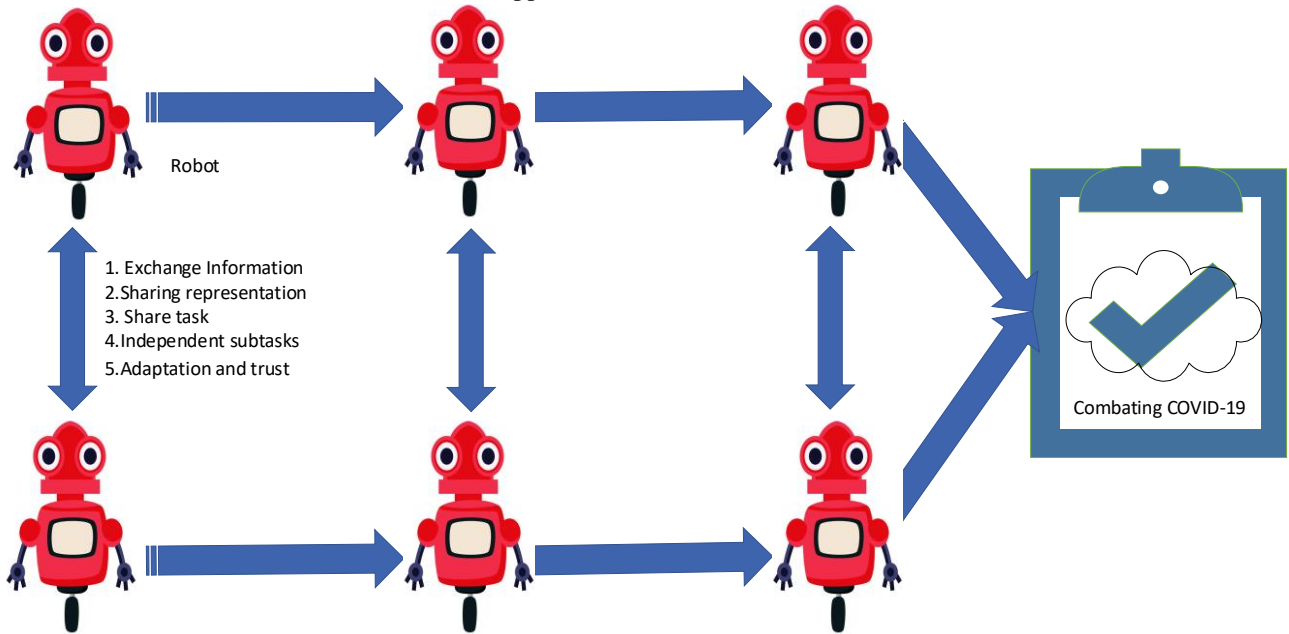

Figure.3 Multi-robot collaboration concept

### 1) MULTI-ROBOT CATEGORIES

Regarding COVID-19 pandemic, robots are categorized into autonomous robots[74, 75], collaborative robots [76], mobile robots, and Telerobots [30], while the function of each category is shown in the figure. 4.

a. **Autonomous robots** - have been worked without human interactions at the time of quarantine. Therefore, autonomous robots can be done without any contact between an infected person and hospital staff.

b. **Collaborative robots -** are known as Cobots, which can decrease healthcare staff and track healthcare staff interactions with an infected person.

c. **Telerobot -** operates remotely and can be used for remote diagnosis, treat an infected person without human interaction. For instance, doctors and nurses can measure the temperature of the infected person of COVID-19 remotely.

d. **Mobile robots-** they work in a hospital and quarantined area to perform autonomous tasks efficiently.

In this article, we focus on Multi-robot collaboration to combat COVID-19, which is presented by more than two robots in the same scene. According to combating COVID-19, multi-robot collaboration can be divided into two categories:

• **Homogenous multi-robot collaboration:** multi-robot has to perform common tasks (delivering, cleaning, spraying disinfection, etc.), which is from the same family for the same purpose. For combating COVID-19 in the hospital, delivery robots are delivering food from the kitchen to rooms. Delivery robots should collaborate to deliver food to all rooms by avoiding collisions and reduce energy consumption. They need to maintain the robustness at a required level, especially with hacked robots, noisy communication channels. Furthermore, multi-robot needs to complete the specified tasks within time and avoid human interaction.

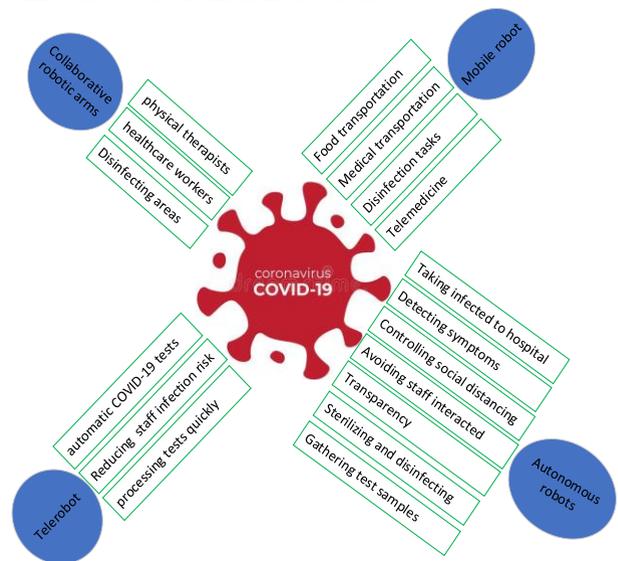

Figure.4 Robots categories for combating COVID-19

• **Heterogeneous multi-robot collaboration:** multi-robot has to perform common tasks (delivering, cleaning, and spraying disinfection, etc.), from different families, shape, and size for a different purpose in the same space. For combating COVID-19 in hospital, heterogeneous multi-robot collaborates with each other in the same space according to their independent subtasks (i.e., cleaning robots and testing robots, spraying disinfection robots, delivery robots, cleaning robots, etc.). Hence, heterogeneous multi-robot can combat COVID-19 in the hospital or quarantine area within time and avoid human interaction.

### 2) FORMALIZATION

We have defined multi-robot collaboration categories in the previous subsection; here, we address more formal models for classification.

*Definition1*. Homogeneous Multi-robot collaboration is defined as a group of robots from the same family to combat COVID-19 in the quarantine area. HMi is a sub-task provided by robot i and presenting as:

$$\begin{pmatrix} HM_1 \\ HM_2 \\ ... \\ HM_i \end{pmatrix} (HM_1 \cap HM_2 \cap ....... \cap HM_i = \in) \qquad (1)$$

*Definition2*. Multi-robot collaboration for multiple sub-tasks to combat COVID-19 in time predefine as set MHMi, where MHMi is the sub-task provided by robots i and present as:

$$\begin{pmatrix} MHM_1 \\ MHM_2 \\ ... \\ MHM_i \end{pmatrix} \qquad (2)$$

*Definition3*. Heterogeneous multi-robot collaboration is defined as HHMj for performing many tasks in the same space to combat COVID-19 in time with energy efficiency, where HHMi is the tasks provide by multi-robot $j^{th}$, and HMi the sub-task provided by Robot $i^{th}$. It represents in the following form:

$$\begin{pmatrix} HHM_1 \\ HHM_2 \\ ...... \\ HHM_j \\ HM_1 \\ HM_2 \\ ..... \\ HM_i \end{pmatrix} \qquad (3)$$

The delivery robot is $D_R$, and smart IoT devices is SIoTD a monitoring robot is $M_R$, speaking robot is $S_R$, cleaning robot $C_R$, spray disinfection robot $SD_R$, a cooking robot $CO_R$, a high body temperature robot $HBT_R$, absent mask robot $AM_R$ . Furthermore, R refers to reducing, $M_{mh}$ refers to the management, and CollabR$_{iH}$ refers to a collaboration in the hospital, while CollabR$_{iQ}$ refers to a collaboration in the quarantine area. While Con$_R$ and Dec$_R$ refer to control and decentralized, respectively. Therefore, several robots are required for different tasks, as shown in table.2. Every robot is created for performing the task, while many robots are required to combat COVID-19. These robots need management and decentralized to collaborate for combat COVID-19, as shown functions below.

$$COVID-19 = avoid_{interaction} + social_{distance} + stay_{home} \qquad (4)$$

$$Stay_{home} = D_R + SIoTD \qquad (5)$$

$$Social_{distance} = D_R + M_R \qquad (6)$$

$$Avoid_{interaction} = R_{iH} + R_{iQ} \qquad (7)$$

$$R_{iH} = C_R + M_R + SD_R + CO_R + MD_R + FD_R + T_R \qquad (8)$$

$$R_{iQ} = E2ED_R + M_R + S_R + HBT_R + AM_R + SD_R \qquad (9)$$

$$CollabR_{iH} = C_R + M_R + SD_R + CO_R + MD_R + FD_R + T_R + Con_R + Dec_R \qquad (10)$$

$$CollabR_{iQ} = E2ED_R + M_R + S_R + HBT_R + AM_R + SD_R + Con_R + Dec_R \qquad (11)$$

Table.2 Robot function for combating COVID-19

| Robot | Combating COVID-19 | Function | Phases | | |
|---|---|---|---|---|---|
| | | | Early diagnosis | Quarantine time | Recovery |
| Thermal image | Interaction between people | Temperature capturing | √ | √ | |
| Disinfection | Prevent workers interaction | Sterilizing contaminated area | | √ | |
| Delivery goods | Reducing hospital visit delivery | Enhancing treatment accessibility | | √ | √ |
| Monitoring | Absent mask, gathering people | Crowd practicing social distance | | √ | √ |
| Announcement | Guiding and delivering information | Broadcasting COVID-19 information | √ | | √ |
| Telerobot | diagnosis, treat an infected person | Sterilize and disinfection | | √ | |

### 3) RESEARCH OBJECTIVES

In this article, we are focused on using a multi-robot for combating COVID-19, which is much more complicated than using only one robot. As multi-robot working together in the same area, the complexity level increases according to the task and area of performing the task. Therefore, multi-robot is sharing the combating COVID-19 and performing different sub-tasks (cleaning, spray disinfection, delivering food and medicine, detecting and tracking specific activity, etc.) collaboratively and efficiently.

## IV. BLOCKCHAIN FOR MULTI-ROBOT COLLABORATION

Blockchain technology is growing exponentially in the robotics field due to allowing robots to conduct a transaction without a central authority. Therefore, it provides data decentralized, accessibility, immutability, irreversibility, and non-repudiation features. Above features make blockchain probably one of the promising and breakthrough technologies [77]. In most cases of robotic swarms, the robot has information about itself or its neighbor robots; however, using blockchain for robotic swarms can provide global information to all robots in the blockchain technology network. Therefore, blockchain can be helpful for robotics applications in the current world. The importance of global information in robotic swarms can lead to appreciating decision making, efficient task performance, easier maintenance. The authors of [21] introduced a blockchain-based protocol for coordinating and controlling of multi-agent in the context of drones. Therefore, blockchain technology enables the drones monitors to log information concerning time, location, resources, and delivers date among swarm robots to avoid collisions, makes data public for all robots.

A swarm of robots serves and monitors a vast number of people in the quarantine area for several purposes in need. Swarm robots need for collaboration in order to share their locations, purpose, and tasks. The authors of [78] explained the development of intelligent multi-robot priority to perform various tasks autonomously. The collaboration of heterogenous robots shares sensitive information among them of the same environment [79]. Furthermore, the tasks of swarm robots need to a collective decision, in which decision-making issues are divided into task allocation and consensus achievement [80]. In the following subsection, we discuss blockchain, robotchain, smart contract, blockchain for homogenous, and heterogeneous multi-robot collaboration.

### A. BLOCKCHAIN

Currently, blockchain technology has been disruptive in academic and industrial domains. The main reason behind using blockchain in the robotic field is to find an optimal model, which is capable of reaching consensus over a reliable multi-robot network. Digital ledger represents blockchain and is designed in a decentralized fashion. It stores information that represents transactions. A smart contract is implemented over blockchain due to it decentralizes data with security. Therefore, accepted transaction into blockchain becomes secure and valid. The authors of [81] summarized the most crucial consensus algorithms characteristics and used. The consensus algorithms include Proof of Work (PoW), Delegated Proof of Stake (DPoS), Proof of Stake (PoS), Practical Byzantine Fault Tolerance (PBFT), PoC, Proof of Burn (PoB), etc. Blockchain classification with features is shown in figure.5. The combination of blockchain and multi-robot makes multi-robot ease of scaling, robustness against failure, operation more secure, autonomous, flexible, and even profitable. To avoid collisions of multi-robot collaboration, blockchain can manage the robots traffic efficiently. Figure.6 illustrates the combination of blockchain and multi-robot to improve the formulating the tasks, distributing decision making, authentication, automating tasks, and action validation. Distributed decision making by voting is challenged by multi-robot collaboration. Blockchain is an efficient solution for distributed decision making. Simultaneously, developed smart contracts in blockchain for multi-robot collaboration can be built for proposing actions and formulating the tasks in bytecode. Then, multi-robot can collaborate with each other to vote for efficient and adaptive action based on both voting and formulating tasks in bytecode. For action validation, robots can check each other such as locations, states, and actions. In the case of robots send incorrect data, the sharding technique can be used to solve the performance of the validator issue and avoid consensus to perform the wrong action. Therefore, the information process is reduced due to the validator coordinate for the shard. For the automated task, the distributed consensus in the blockchain is useful for dispatching, assigning, and executing tasks between multi-robot collaboration. A smart contract can improve the authentication of multi-robot collaboration.

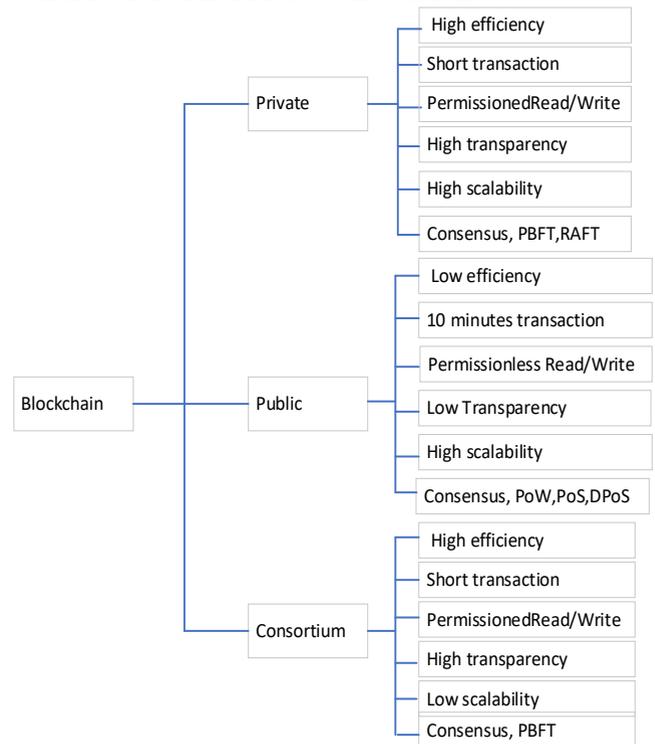

Figure.5 Blockchain types and features

a. Blockchain for homogenous multi-robot collaboration
Blockchain plays a vital role in managing robot traffic efficiently with high accuracy. Furthermore, it helps to secure the operations of robots, ensure robots stay on track (no crash, no harm, no injuring), have not diverged private information. The authors of [21] introduced the blockchain technology-based efficient protocol for the coordination of multi-robot in the context of robots. The importance of using blockchain technology was to update the real-time location of robots while delivering goods and medical supplies. For combating COVID-19, avoid a collision is required. Due to public data in the blockchain, each robot can access other robots status, as shown in figure.7. Multi-robot can update location, status, environment, and event in blockchain as well as share decision making collaboratively in real-time with energy efficiency.

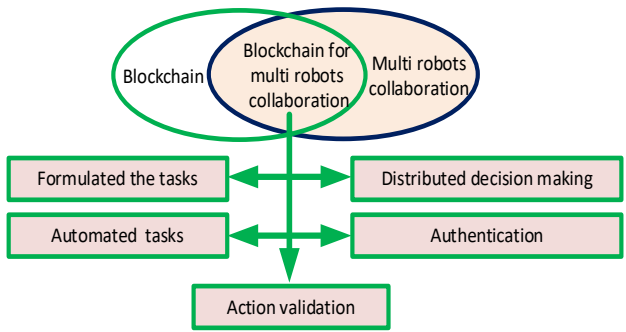

Figure.6 Blockchain for multi-robot collaboration

b. Blockchain for heterogeneous multi-robot collaboration
Many robots are used for combating COVID-19 in hospitals, including autonomous spray disinfectant robots, medical tests robots, hospital robots, ultraviolet disinfection (UVD) robots, etc. Hospital robots are developed to detect body temperature and maintain access control [82]. An autonomous spray disinfection robot is used in multiple forms to perform tasks in every corner of the quarantine area. UVD robot can move indoor (i.e., around patient rooms) and outdoor (i.e., school and hospital campus) autonomously, shining the right amount of UV-C light needed to kill specific bacteria viruses in the patient room [83]. Furthermore, robots deployed in order to deliver food inside the hotel in Hangzhou, China, in order to avoid stuff service dangerous [84]. Robots are delivering items to people who are quarantined at home.

Robot manufacturers join fighting COVID-19 by creating autonomous models during navigation and assisting human cleaning surfaces where virus particles can spread among an infected population. Robots can deploy for disinfection, delivering medications/food, autonomous spray disinfectant, and cleaning the floors [85], detect body temperature in public areas. Developed disinfected robots help to detect high risk, high touch areas, and clean continuously. Measuring robots is used to measure the size of an obstacle and then relay information about the obstacle to manipulating robots to make a decision accordingly. Here, blockchain stores all historical information about proposed robots for performing the task. Furthermore, blockchain-enables multi-robot to provide security in the presence of malicious robots, exclude and detect byzantine robots from contributing to consensus in historical action in the distributed ledger [17].

### B. ROBOTCHAIN
Robotchain system is proposed based on the blockchain to register the events of robots in a secure and trustworthy manner [86]. The architecture of blockchain-based Robotchain is shown in Figure.8. Robotchain includes robots, a control unit, and a decentralized blockchain network. The control unit includes a monitoring robot, blockchain clients, and a control robot. Multi-robot can form a robot cluster in order to perform combating Covid-19 either in the hospital or outdoor. The robot body consists of many sensors to gather biological data.

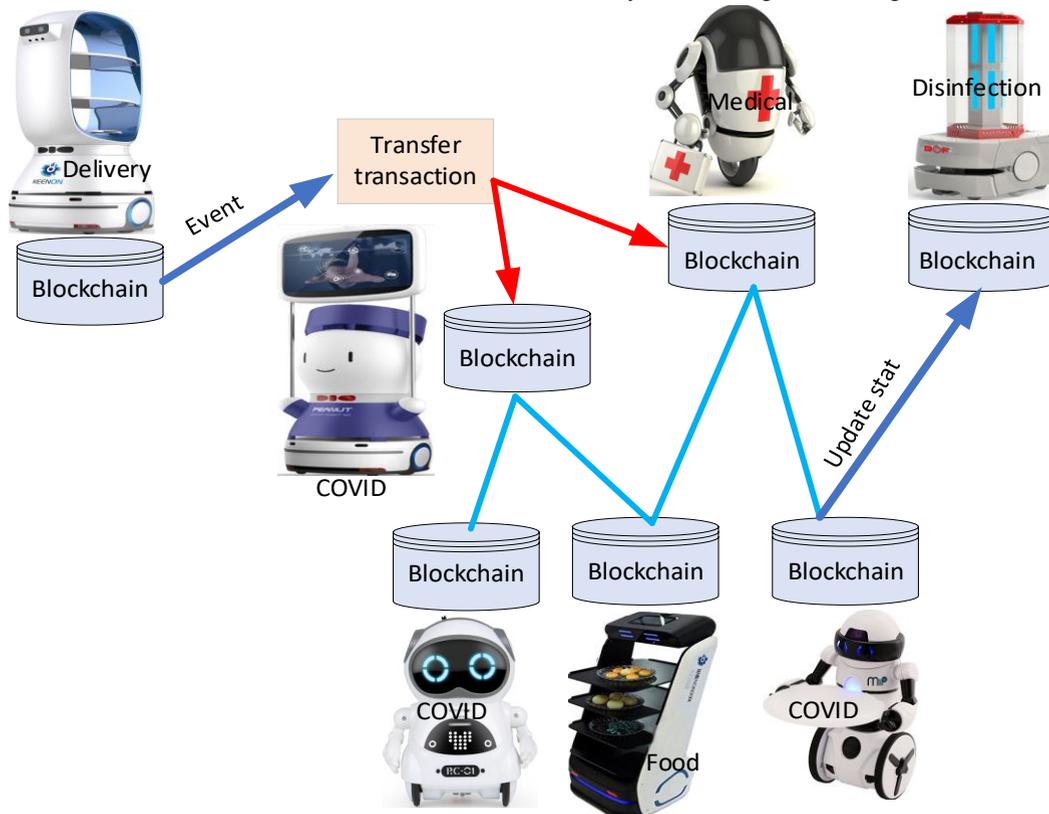

Figure.7 Robot access all robots state in blockchain

Furthermore, multi-robot requires connecting with the control unit to send gathered data, robot event status, and receiving commands. The controlling unit can interact with a multi-robot for gathering data. It is responsible for receiving gathered data, sending commands, and controlling the robot's behavior. It works to send both hashed data and original data to the blockchain network. A decentralized blockchain network can be used for resilience and data validation. It is used for a few purposed, i.e., data gathered from robots, commands from the control unit, and integrity protection. Therefore, blockchain helps multi-robot collaboration in the quarantine area to collect data from robots, respond from the control system, store data in a distributed fashion to ensure stability, share data and decentralize to perform decision-making real-time.

Robotchain main characteristics are including easy verification, self-amending, consensus algorithms. It is the support of easy verification of smart contracts. Simultaneously, self-amending allows the changes to be performed by voting the chain on the blockchain. Furthermore, the process self-amended is helpful when the change in protocol is required for minor changes. Consensus algorithms are essential to deal with big data, where time and energy are required to validate transactions efficiently. RobotChain is an approach in which applications can be created on the top using Oracles or by inserting smart contracts with different actions on the blockchain.

### C. SMART CONTRACTS

A smart contract is a digital and automatic transaction protocol (rules of contract) placed in a specific address in blockchain to perform processes accordingly. The protocol is deployed in blockchain [87] and can self-execute as the agreement met. A smart contract can perform calculations, storage information, transfer, etc., automatically [88]. Furthermore, it can support polymorphism and inheritance [89]. Data authorization rules, functions, and processes are embedded in smart contracts. Therefore, decentralized can be managed via smart contracts, which can reduce the managing cost processes significantly. Evaluation and development are types of smart contracts. Evaluation is including code analysis and performance, while the performance of development can be under development.

Ethereum addressed Solidity [90] as a language (including code instructions and event state (data) like the initial, intermediate, and also final) in order to implement smart contracts. Furthermore, the authors of [90] introduced smart contracts as part of the Ethereum virtual machine. Ethereum virtual machine is an isolated environment while accessing data among smart contracts is limited. The transactions contain outputs of smart contract code. Transactions code execute within the Ethereum virtual machine.

A smart contract is capable of actions, including data collection, processing, and adopting specific solutions. A smart contract on the blockchain; the contract is formulated in the programming language. Then, the smart contract transfers to a blockchain, which will be automatically self-executed as the agreement are reached. Once the specific event takes place, and the transaction reaches the smart contract, the blockchain distributed virtual machine executes the code of the program. A new party can join a smart contract and initiate automatic execution via meeting specific historical conditions.

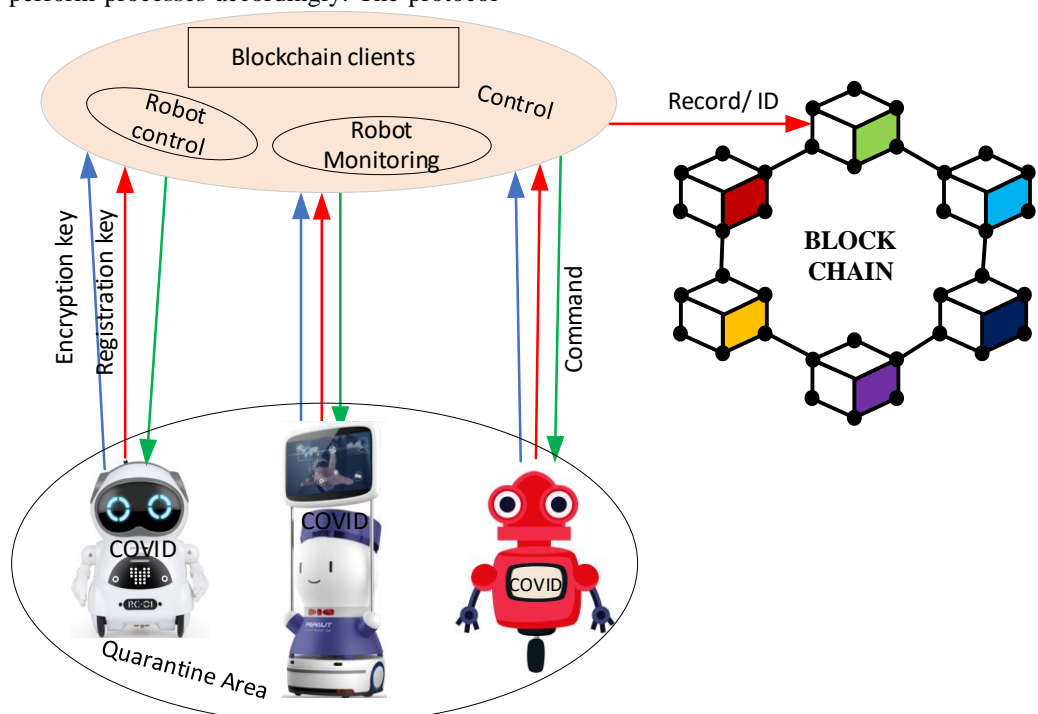

Figure.8 Robotchain communication architecture and operation

## V. PROPOSED FRAMEWORK SOLUTIONS

Regarding combat COVID-19, we focus on identifying the challenges in multi-robot collaboration. Then, we discuss the tactical solutions offered by using blockchain technology for identified challenges. Thus, we address proposed framework solutions by using blockchain for multi-robot collaboration challenges in combating COVID-19. We mainly focus on controlling and decentralized multi-robot collaboration. Then, we address applications of multi-robot collaboration in the monitoring quarantine areas, outdoor E2E delivery system, E2E delivery system in hospitals.

### 1) CONTROLLING MULTI-ROBOT TO COMBAT COVID

The proposed network architecture is by developing novel techniques for controlling robots in multi-robot collaboration. Blockchain technology, as the decentralized ledger, is used in the development of proposed network architecture for multi-robot collaboration in the quarantine area. Blockchain is used to store information about the robot sensing environment (sensors, cameras). Smart contracts are used to define the robot state and store information in order to control robots (i.e., action, behavior, navigation, resources, etc.) of multi-robot collaboration. The information received from robot shares with other robots via a smart contract. Smart contracts sent information to edge intelligent computing in order to change the required robot action and behavior locally for performing specific tasks effectively and efficiently. Edge intelligent computing unit handles the interaction between multi-robot intelligently, while the monitoring unit can access the blockchain and follow robots until performing a specific task according to adopting efficient analysis methods. Figure 9 illustrates the possibility to control robots in the team with the help of smart contracts. It shows the composition of the proposed architecture, such as robots and ledgers. Robotchain represents the ledger, which contains a smart contract.

In figure 9, robot 1 detects something valuable/essential data, AI in edge computing analysis data. Then, robot 1 sends important information (block, hash, and ID) to robotchain. Robot 2 to Robot N can access the information sent by robot 1 via smart contract as log and information. The monitoring unit can access smart contracts and analysis. The monitoring unit can add block (hash and ID). Finally, smart contracts in robotchain send comments for controlling all robots in multi-robot collaboration and change robot behavior according to smart contracts comments.

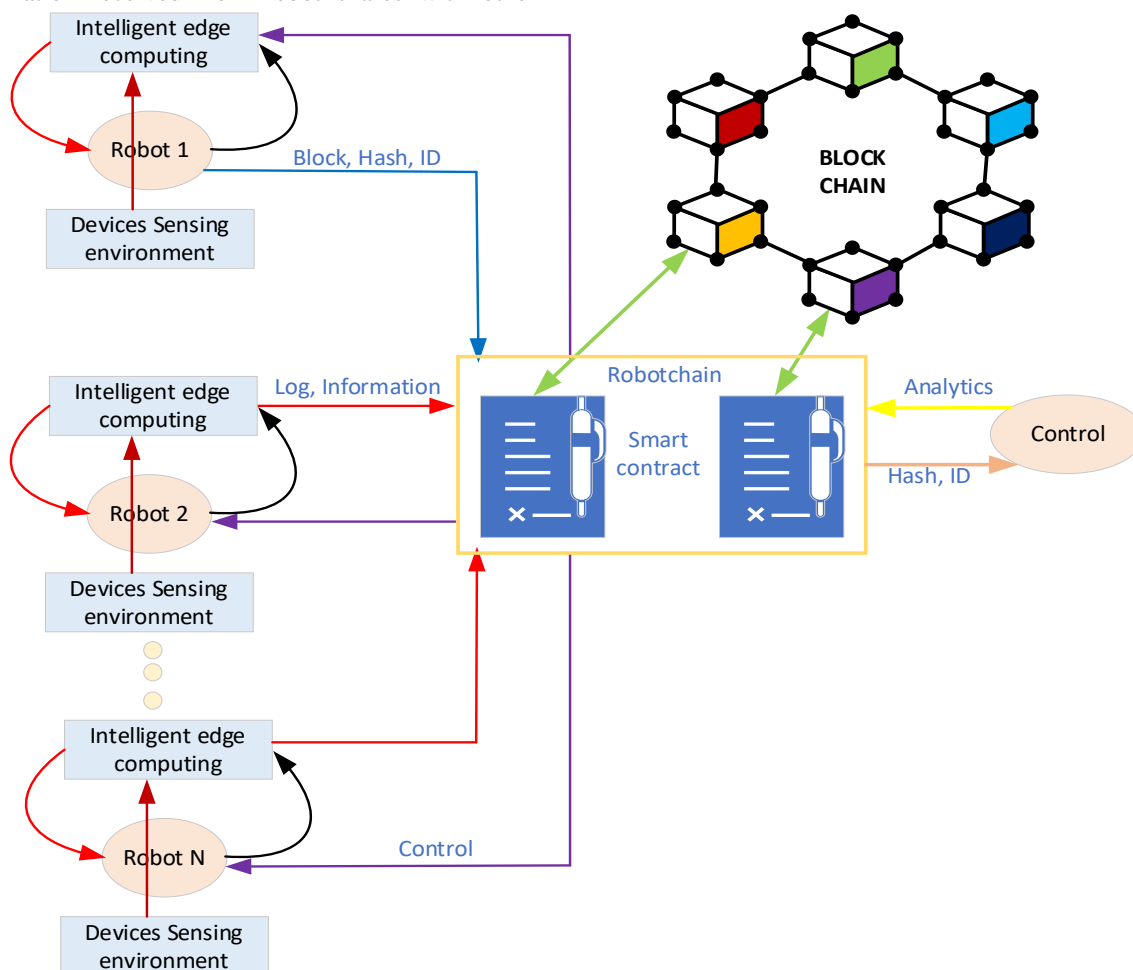

Figure.9. Multi-robot control for combating COVID-19

Blockchain network technology helps to update the locations of each robot on the distributed ledger in real-time. Each robot can access the real-time location of all robots in the blockchain network to avoid collisions. For instance, all robots are used for a different purpose in the quarantine area (i.e., delivering the medical supplies and goods, monitoring and detecting the people who are breaking the rules, automating spray disinfectants, and identifying the infected person of COVID-19). All robots update (speed, location, resources, delivery time and date, and ID) to blockchain network technology in the real-time and authenticated manner.

The proposed framework solution of combining blockchain and multi-robot can control homogenous or heterogeneous multi-robot with avoiding collisions while performing tasks in the quarantine area. Therefore, each robot has several sensors for different needs. For instance, the robot has a stop sensor in order to stop robots. Robotchain is a decentralized ledger that is used to store robot history events, status in high securing. Furthermore, the robotchain is capable of handling a large number of transactions that are produced from multi-robot per second. Moreover, controlling robots in multi-robot collaboration does not need human intervention. The reason behind this is that smart contract availability can select suitable action automatically based on gathered data from the quarantine area. It notes that there is no human interaction in the operation area of multi-robot collaboration. Moreover, it observes that controlling multi-robot collaboration by blockchain can be applicable for real applications such as fighting COVID-19 and future pandemics. Furthermore, controlling robots becomes easy to adapt and maintain the new robot task/ action since a new smart contract handles the new requirement. Therefore, joining any robot cannot affect the combating of COVID-19 and future pandemics system.

### 2) DECENTRALIZED MULTI-ROBOT COLLABORATION

Multi-robot can assist in many purposes in combating COVID-19, such as monitoring, spray disinfection, delivering, detecting, and identifying infected cases of COVID-19. However, a managing multi-robot network needs for protocols and efficient techniques to perform their complex tasks efficiently in real-time. Furthermore, multi-robot needs collaboration to share their locations, flying speed, decision making, battery status, purposes, and tasks. Multi-robot collaboration needs to reach the common goal based on joint intention and planning, but decision-making during action planning is challenged. With the support of the IoT, Cloud/Edge computing, and AI, multi-robot has become adaptive, collaborative, and connected, thereby enabling complex interactions amongst decentralized robots.

Homogenous and heterogeneous multi-robot collaboration networks suffer from a single point of failure, whereas a decentralized robot network suffers from lacking global knowledge. For instance, robot malfunction is one common problem in the collaboration of multi-robot. The malfunctioning robot in a centralized network cause destroying the robots function network due to the dependency of robots. The dependency makes the center are unable to reach to any robot data if there is a malfunctioning robot happened. Multi-robot collaboration may be partitioned dynamically, and there are no communication channels between partitions due to communication challenges or navigational. Network partitioning can disturb multi-robot collaboration operation and decrease the effectiveness of combating COVID-19 outbreak. Different network portioning in heterogeneous multi-robot collaboration may be in different actions, states, and needs, leading to conflicting. Here, managing portioning tolerance and adapting network partitions become an essential requirement for multi-robot collaboration functionality to combat COVID-19. Therefore, we introduced a decentralized ledger platform with corresponding protocols and algorithms to enable heterogeneous multi-robot collaboration with a tolerance of network partition and malfunctioning or byzantine robots in the hospital or in the quarantine area.

Many works have proposed to use blockchain technology to facilitate multi-robot collaboration and management in literature. Blockchain adapted to support many events quickly to deal with many interacting robots in the network. Furthermore, blockchain can improve transaction speed and how a multi-robot change behavior to combat COVID-19. Blockchain is an essential technology for improving the decentralization issue, in which swarm robots are suffering. It is of paramount importance to use blockchain technology for sharing data between robots at a high level of authentication with energy efficiency and high accuracy. Sharing smart contracts in the blockchain network between robots allows each robot to read other robot's data. Therefore, multi-robot can reach to consensus collaboratively to combat COVID-19 effectively in real-time.

Planning actions and agreements are related to transactions stored in the blockchain, and therefore, no need for training and learning in case of new robot joined or failed. A new robot can download the ledger from a smart contract and then start planning action accordingly. However, a malfunctioning robot can be replaced by another one quickly. Furthermore, the robot can borrow missing information from neighbors one due to a malfunctioning device. Blockchain technology becomes more critical due to support the decentralized and share data in real-time among multi-robot collaboration. With blockchain, robots play a crucial role in controlling public health emergencies such as combating COVID-19 outbreak.

The consensus algorithm improves multi-robot collaboration, with a focus on network partitioning and byzantine members. Fault-tolerant consensus algorithms are forced to make trade-offs within a triangle formed by the number of robots. Therefore, multi-robot reaches the consensus, the communication overhead of the consensus protocol, and the latency required to reach the finality.

Furthermore, sharding is a scheme to increase blockchain scalability, so that the transaction processing and world state are limited in the sharing range, as shown in figure.10. As the sharding architecture matches well with the community feature of robots, the existing sharding scheme can be improved to accommodate fault tolerance. The consensus algorithms with dynamic sharding techniques based on location, task types, community (membership), flight speed, population, etc., are used to support network partitioning with malfunctioning robots. Consensus algorithms enable robots collaboration with a tolerance of network partition and malfunctioning robots with improving the decentralization of multi-robot at low cost.

Furthermore, blockchain technology has not been used until now for developing the decentralized ledger platform with a consensus algorithm to improve multi-robot collaboration. Secondly, scalability is also a problem in joining a new robot to blockchain network technology due to limit sharing rang. Therefore, the integrated dynamic sharding techniques and consensus algorithms in the blockchain is supporting the scalability of robot collaboration. Flexible and more autonomous solutions to robot decision-making in distributed systems must tackle the new wave of challenges. For example, in performing every sub-task, each robot needs an agreement from other robots, as shown in figure 10. With the help of blockchain, decision-making stores in the block, and shares for public robots in the group.

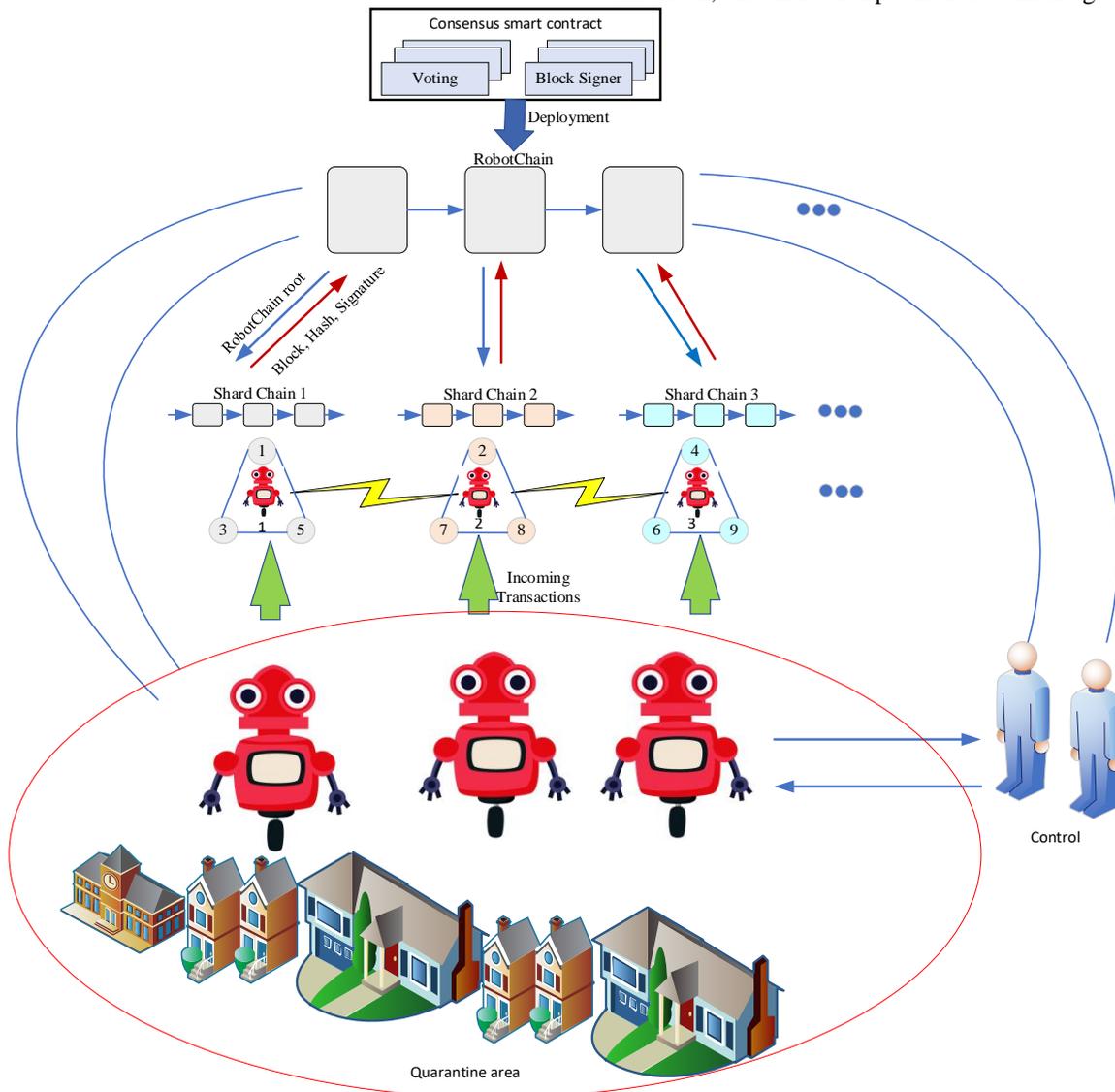

Figure.10. Consensus algorithm with sharding technique in blockchain for multi drone

Briefly, taking advantage of both the sharding technique and consensus algorithms can improve network partitioning, robot malfunctioning, and scalability of the blockchain network. Therefore, it is expected to develop multi-robot collaboration, more realistically, and dynamically. Consequently, the response time for taking action and decision making with the help of blockchain technology for multi-robot collaboration can be reduced with better performance and energy efficiency.

a) Consensus algorithm

Consensus algorithms in a decentralized blockchain network allow robots to reach an agreement and transactions whenever it is required. Figure.11 shows suitable consensus algorithms for the development of blockchain types. Consensus algorithms allow robots in the blockchain network to trust with all robots. The consensus algorithm plays a vital role in managing the collaboration in multi-robot [91]. In [81], the consensus algorithms are discussed with details. Consensus algorithms in blockchain can provide solutions for decentralized, secure, and scalable in multi-robot collaboration [92]. Fault-tolerant consensus algorithms are forced to make trade-off within a triangle formed by the number of robots. Figure 11 illustrates the consensus algorithms (i.e., PoW, PoS, and PBFT) benefits and limitations.

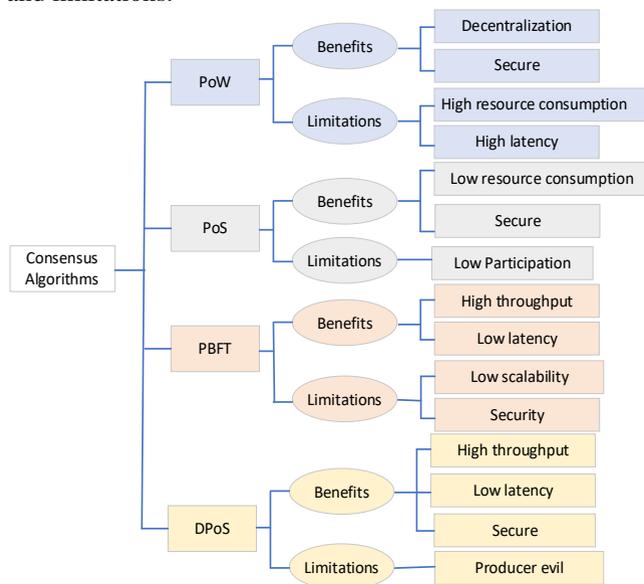

Figure.11 Consensus algorithms benefit and applications

b) Sharding technique

The enormous computational resources are the reason behind the challenges of PoW salability [93]. Furthermore, PBFT is suffering from scalability due to communication costs and broadcasting messages [94]. The authors of [95] addressed scalability limitations of blockchain and introduced the technique in which the network is partitioned into small subnetworks. The implementation of the sharding protocol was for permissionless blockchain. Sharding can tolerate byzantine robots predefined fraction in the network. Then authors of [96] proposed sharding in the blockchains, in which Merkle Patricia trees or Merklix trees [97] were used to different shards state to global state. Currently, the most significant effort is being given to decentralized development, scalable, and permissionless secure blockchain, which designs of Ethereum 2.0 [98]. Therefore, considerable computational resources are not required for mining [99]. Ethereum 2.0 has embraced PoS as a consensus algorithm, while PoW uses to secure the network and sharding for scalability. In order to deploy smart contracts in the blockchain network with shards, two techniques can be applied. First, the shard synchronization technique can allow data flow between shards. Second, the contract is executed within every single shard. The Directed Acyclic Graph (DAG) [100, 101] and membership [58] based approaches will be considered to integrate with different sharding scheme. It notes that the proposed framework solution for combating COVID-19 can be established a much more realistic collaboration that could avoid failure tolerance and network partitioning.

## VI. APPLICATION DOMAINS

### 1) MONITORING QUARANTINE AREA

The robot body contains many smart devices for different needs and purposes. Smart devices are used to capture environmental data. For instance, for combating COVID-19, robots are equipped high-resolution camera to capture images of absence mask, speak announce for social distance and sperate people gathering, thermal camera for detecting infected persons of COVID-19 earlier. With the help of AI, robots can recognize the activities according to the use purpose (detecting infected person, detecting absence mask, detecting gathering people, recognizing social distance between people). The blockchain serves as a ledger to store information in different communication protocols, share stored information to all robots in a decentralized blockchain network, as shown in figure.12.

Smart contracts are used to interact between heterogeneous robots and control robots in the quarantine area. Edge intelligent computing unit handles the interaction between robots in the blockchain. An intelligent algorithm is prepared for activity recognition (absence mask, high body temperature, social distance), which can extract and classify the unique features of a person. Then, a smart contract will identify the needs actions to be taken by the monitoring units (absence mask, high body temperature, social distance). Smart contracts send information to edge intelligent computing in order to change robot actions and behavior for performing specific tasks effectively and efficiently. In this context, the information about how many infected people is stored in the blockchain. If the number of infected people changes, the smart contract automatically determines and updates infected people. Furthermore, robots are coupled with a computing unit that can handle the interaction with robots in the blockchain, as shown in figure.12.

In detecting absence mask, the robot detects absence mask person and social distance from long distance. Then, the robot produces an alarm immediately and registers the event automatically in the blockchain. The mask detection unit should be prepared to take action against the detected person in the case alarm still updates and registers in the blockchain (robot ID, detected person location, movement, direction, etc.).

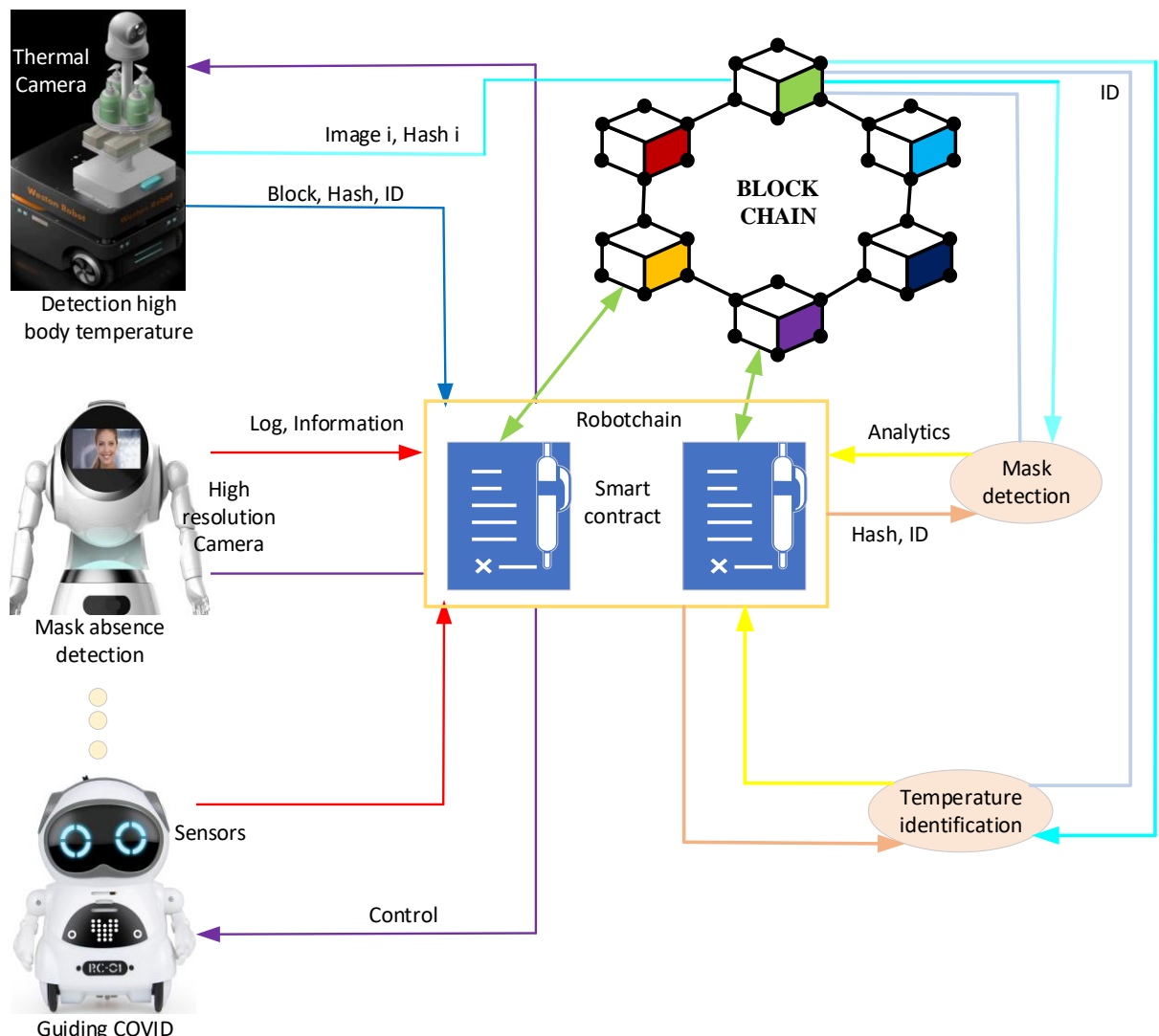

Figure.12. The architecture of multi-drone collaboration for high body temperature and mask absence

In figure 12, the high-temperature robot detects an infected person of COVID-19 from a long distance by using a thermal camera. Robot analysis the gathered data about the capture thermal image intelligently. A high body temperature robot sends important information that extracts from the captured image (block, hash, and ID) to robotchain. High body temperature robot registers the infected person registers in the blockchain (robot ID, infected person location, movement, direction, etc.). All robots in the blockchain network will be informed via a smart contract. These robots can access the information sent by high-temperature robots via smart contracts as log and information.

Therefore, the guide/ announced robot produces an alert immediately (detecting infected person) and guides the infected person and people nearby. Furthermore, spray disinfection comes closer to the infected person for autonomous spray disinfectant, while the delivery robot provides the required medicine and mask in case an infected person is an absent mask. A self-driving robot takes the infected person to the nearest hospital to ensure that the detected person is infected with COVID-19. Also, the monitoring unit can access smart contracts and analysis intelligently the information sent by high-temperature robots. The monitoring unit can add block (hash and ID). Finally, smart contracts in robotchain send comments for controlling all robots in multi-robot collaboration and change robot behavior according to smart contracts comments.

The procedure of monitoring multi-robot collaboration to combat COVID-19 is shown in figure.13. It also shows how blockchain helps to control and decentralized multi-robot collaboration.

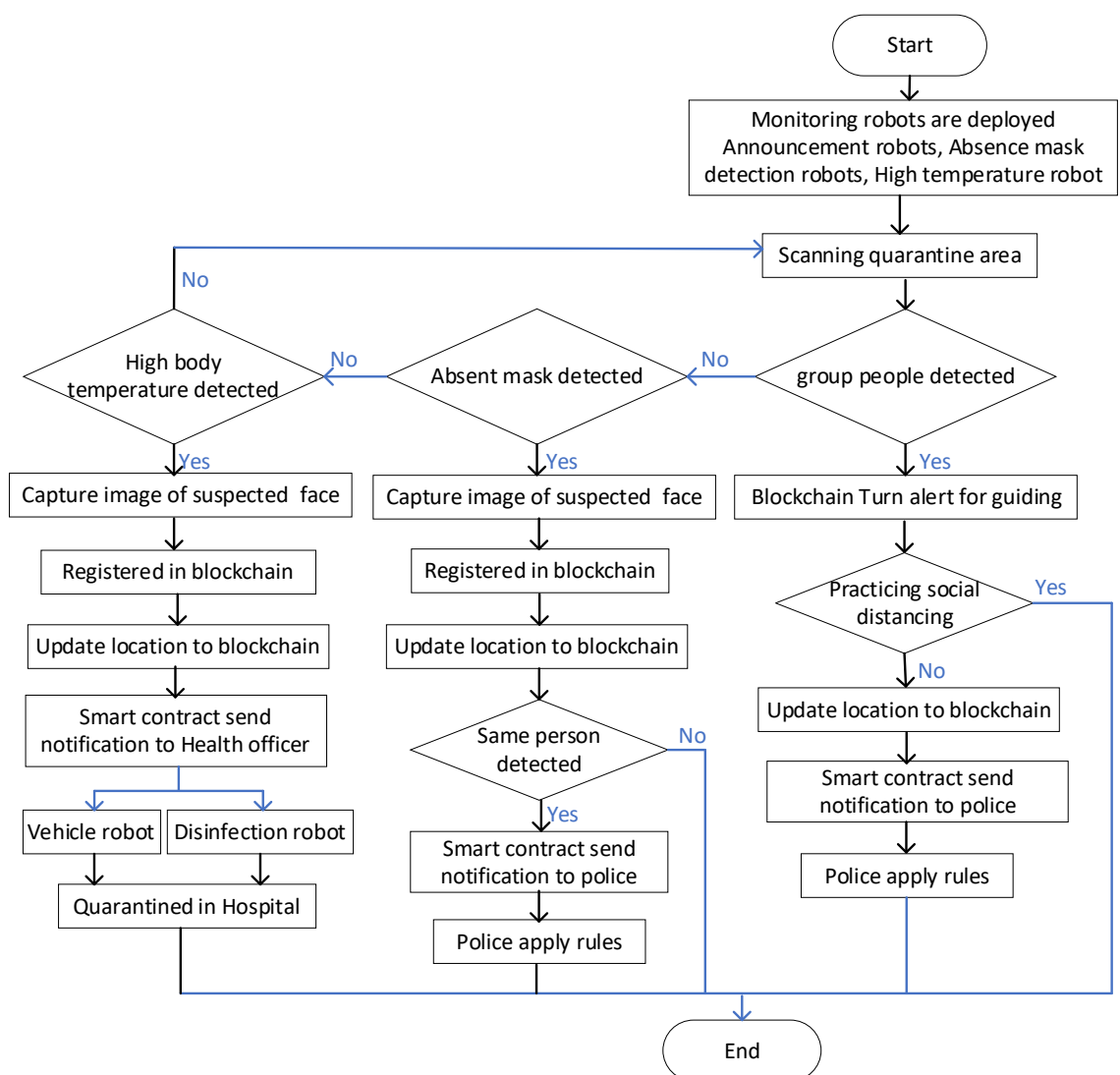

Figure.13 Algorithm of monitoring multi-robot collaboration to combat COVID-19

## 2) QUARANTINE E2E DELIVERY SYSTEM

Delivery robots can serve people in the quarantine area because robots can transport medicines and goods. In this case, we focus on delivery robots for medicine and food to the quarantine area. In order to implement the use cases, reliable and fast robots, network connectivity is required, such as 5G. Furthermore, mobile network connectivity is preferable to provide high quality, secure connectivity to improve the security and safety network of robot communications. Moreover, a security attack on robot communication networks is needed in order to control steal or spoof attack vectors. Attack vectors are including monitoring (i.e., monitoring data from the center unit), control (i.e., getting data from and to the center unit), and data acquisition (i.e., location, speed, and sensing environment). To avoid attack vectors, a combination of blockchain and delivery robots is a suitable solution over 5G. Therefore, 5G is offering various benefits to the combination of delivery robots and blockchain. The benefits include high data reliability in goods delivery, faster access to data due to decentralized characteristic, high availability of data due to a distributed ledger, network fault-tolerant due to each robot has a copy of the ledger, the resistance from data modification attack, and the new robot can join blockchain after all robots verification.

Figure 14 illustrates the proposed framework solution for delivery robots in the quarantine area. The proposed framework network architecture consists of three layers, including the donate layer, blockchain layer, received layer. The received layer requests medicine and food from the donating layer through the blockchain network by Ethereum. Donated prepared the order with encrypted data from donated (path, receiver, and robot ID and location). Smart contact in blockchain verifies and executes the status condition between donated and receiver, then stores the verification in the blockchain network. Therefore, the data in the blockchain is secure, reliable, and traceable. In such a delivery system, a quarantined person makes an order for delivery (online).

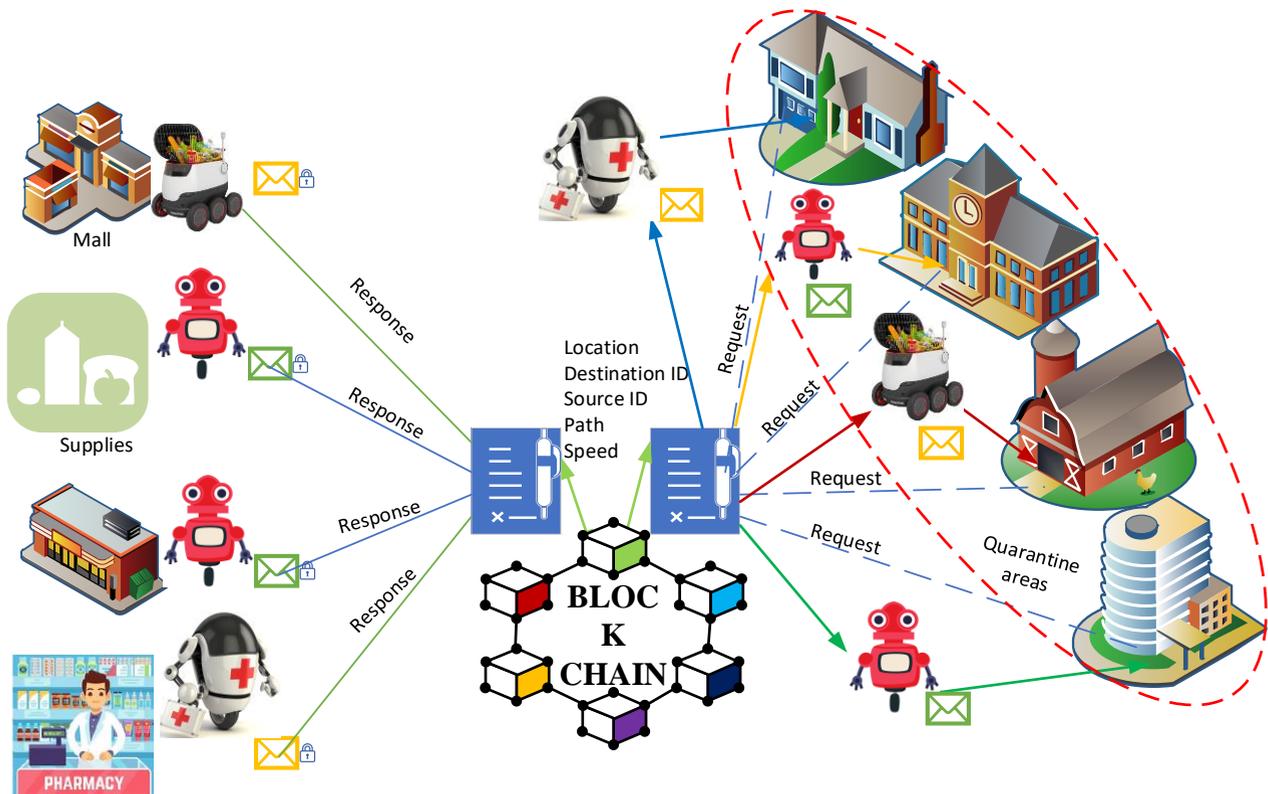

Figure.14. Delivery of food and medicine in blockchain envisioned robots networks

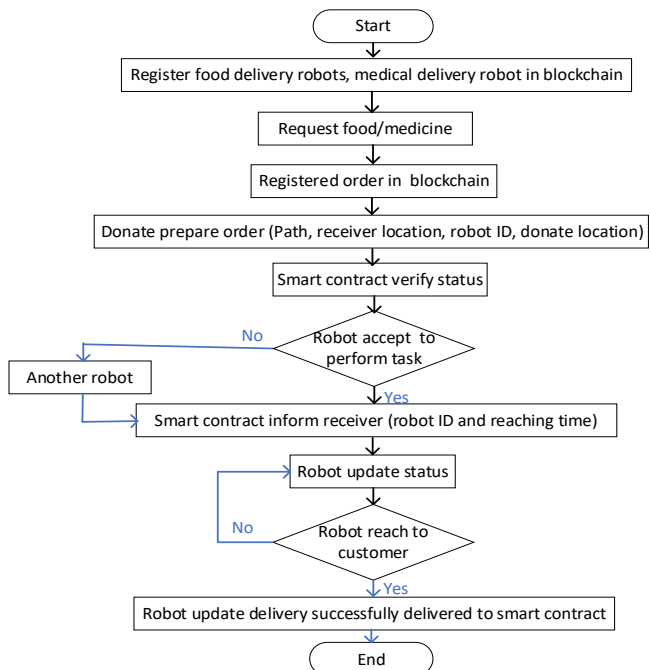

Figure.15 Algorithm of multi-robot collaboration for quarantine E2E delivery system

The smart contract will be generated automatically with sufficient data about (the order purpose, person location, donated location, trajectory, etc.) and transfer the smart contract to the blockchain. Here, the required delivery robot will accept to perform this contract. Then, a person is quarantined to inform about assigned robots, the required time for delivery, and registered trajectory. As soon as the donated made order ready, the assigned robot starts performing the missing and updated location simultaneously to the smart contracts. Finally, the robot notified both people and denoted that order has been delivered successfully via updating dispatching and returning to the source station information in the blockchain. Figure 15 illustrates the procedure of the outdoor E2E delivery system and blockchain function during multi-robot collaboration to deliver goods during COVID-19.

### 3) HOSPITAL E2E DELIVERY SYSTEM

In the hospital, E2E delivery system is critical to avoid interaction between hospital staff and infected persons of COVID-19. A delivery robot can deliver goods or foods to the patients inside the hospital, as shown in figure.16. Figure.16 shows many robots are working together in the same space in order to combat COVID-19. Several infected patients are admitted to rooms. Disinfection cleaning, nurse, delivering food, and medicine, and monitoring robots are used to serve the infected patients of COVID-19. In this scenario, blockchain technology helps in collaborating with heterogenous multi-robots in order to combat COVID-19 collaboratively, intelligently, and efficiently in a decentralized fashion without collusion.

In E2E delivery system, delivery robots update location, speed etc. into the blockchain, while the blockchain shares the updated information with all robots, infected persons, etc. via smart contract. A specific robot to deliver particular tasks, reaches the infected location, package registers in blockchain, and status updated (package has delivered successfully). Previously, robots in the hospital, Wuhan, China, are used for delivering food to the patient room via speaking at the room door.

For instance, testing robot/ nurse robots can update infected person status to the blockchain, doctors access smart contracts and suggest suitable medicine for current status to the smart contracts. Pharmacia checks the smart contracts and prepared essential medicine according to doctor suggestions. Then, the delivery robot comes to deliver suggested medicine from Pharmacy to an infected person, including an explanation of how to use the suggested medicine. An infected person opens the door and takes medicine.

Therefore, the interaction between doctor, nurse, and hospital staff is not required. Therefore, it notes that delivery robots collaborate with other robots in hospital plays a critical role in combating COVID-19 outbreak effectively and efficiently in a high-security. In smart hospitals, smart IoT devices placed at the door can join as new nodes in the blockchain. Then, the infected person's room door can open automatically as soon as the delivery robot reached the door. Blockchain is used to store robot updated data to ensure the door opens for a specific robot. As the robot reached near to the patient's door, the blockchain updates the location, speed and shares it with the door. Therefore, IoT devices in the door of the supposed room enable the door to identify and verify the blockchain-based automatic share of the updated location of robots in smart contracts to open the door and let the robot enter the room and close the door when robots go out. Here, the blockchain technology network enables the interaction between smart IoT devices placed at the door and robots collaboratively to deliver things to a patient bed. Therefore, the collaboration of robots and IoT devices can help robots to reach to an infected person in a critical situation due to COVID-19. For E2E delivery system in the hospital, figure 17 shows the procedures and functions of several robots to combat COVID-19 with the helped of blockchain technology.

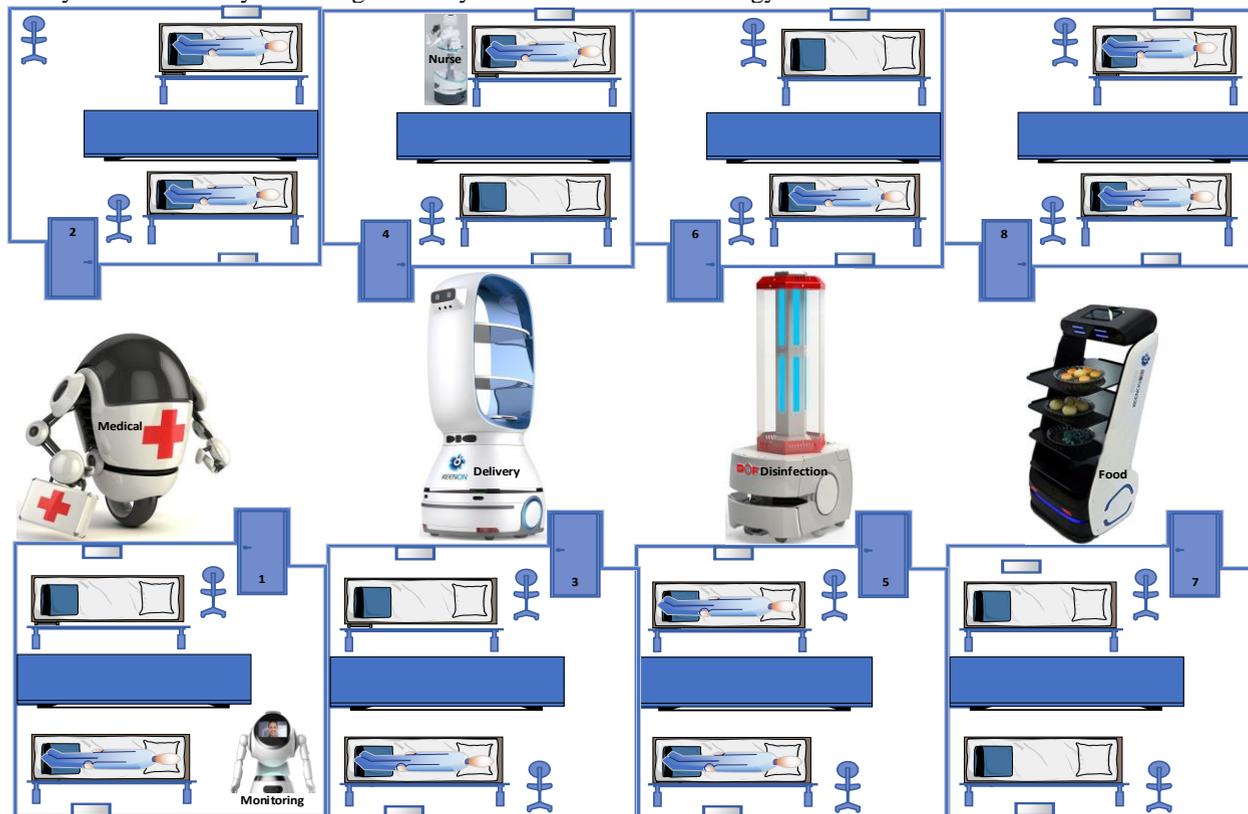

Figure.16 Multi-robot collaboration to combat COVID-19 in hospital

Briefly, the summary of study cases and the proposed framework is shown in table.3. It describes the uses of robots for combating COVID-19 outbreak either in hospitals or in the quarantine area. The primary purpose of robots for combating COVID-19 is reducing people's interaction by monitoring, broadcasting, replacing the human, spray disinfection, cleaning, delivering goods, and medical supplies. However, several problems may occur during robot task performance, such as collusion, navigation, joining a new robot to the team, malfunctioning, and E2E delivery system. Blockchain technology network represents the critical technology in offering an appreciated solution for robot collaboration issues for combating COVID-19 outbreak. Algorithms and techniques are also discussed in order to improve blockchain technology in robot collaboration, such as consensus algorithms and sharding techniques.

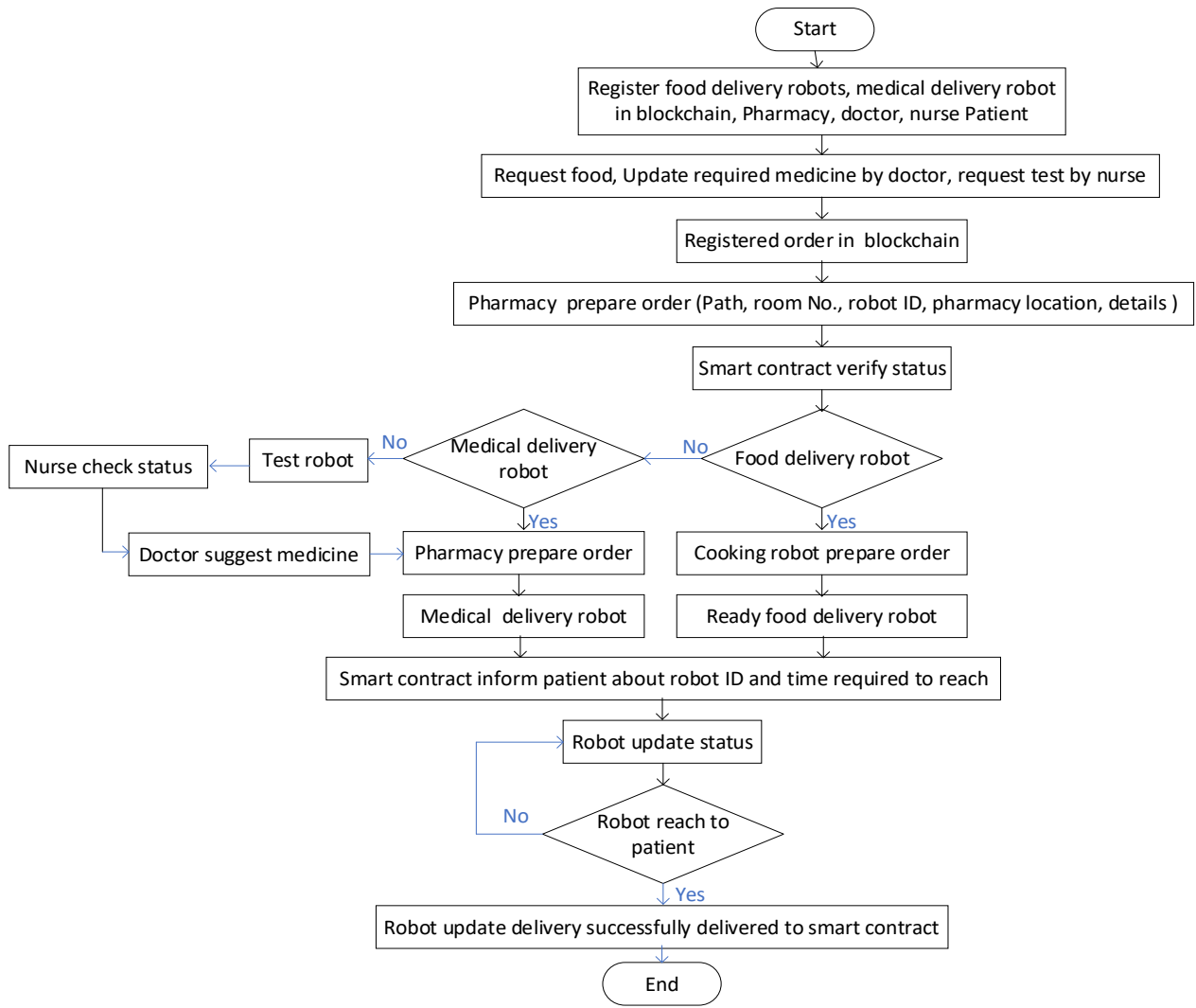

Figure.17 Algorithm of multi-robot collaboration for hospital E2E delivery system

Table.3 Proposed framework solution summary for combating COVID-19 outbreak

| Problem | Proposed solution | Activities required | Issues may be faced | Blockchain for issues solutions |
|---|---|---|---|---|
| Interaction between staff and patients in the hospital.<br><br>Delivering services. | Heterogenous and homogenous multi-robot collaboration to combat COVID-19 outbreak. | • Avoiding stuff and patient interaction by:<br>• Delivering goods and medical supplies<br>• Spray disinfection<br>• Replacing stuff in some jobs, such as cleaning and test patient body temperature. | ○ Malfunctioning<br>○ Centralization<br>○ Supervision<br>○ Scalability | Blockchain decentralized characteristic solves network partitioning, malfunctioning, and improves the scalability in multi-robot collaboration. |
| | | | ○ Navigation<br>○ Collision | Blockchain shares information of each robot, controlling multi-robot collaboration becomes easier while performing several tasks. |
| | | | ○ Navigation | Blockchain keeps updating the location and speed of each robot, so multi-robot collaboration delivers goods, foods, medical supplies. |
| Interaction between people in the quarantine area.<br><br>Monitoring people<br><br>Delivering services | Heterogenous and homogenous multi-robot collaboration to combat COVID-19 outbreak in the quarantine area. | • Detecting high body temperature and the absence of a mask.<br>• Delivering goods and medical supplies.<br>• Spray disinfection.<br>• Broadcasting | ○ Malfunctioning<br>○ Centralization<br>○ Supervision | Blockchain decentralized solves multi-robot collaboration to improve supervising and malfunctioning robots. |
| | | | ○ Collision<br>○ Navigation | Blockchain shares robot location in real-time to control navigation and avoid collision during task performance of multi-robot collaboration. |
| | | | ○ Collision<br>○ Navigation<br>○ Trajectory planning | Blockchain shares each robot information with others for delivering goods, foods, medical supplies. |
| | | | ○ A robot cannot reach inside the customer room | Blockchain shares robot information with IoT devices in the smart hospital, so robots can automatically deliver the patient bed or desks. |

## VII. DISCUSSION

Currently, robots are a promising solution for avoiding interaction between people during COVID-19 outbreak by delivering goods and medical supplies, spray disinfection, cleaning, monitoring, detecting high body temperature, or mask absence, broadcasting information, etc. In this section, we discuss how blockchain can provide numerous uses in multi-robot collaboration for combating COVID-19.

### 1. CHALLENGES

The objectives of this paper are to define the proposed framework solutions for multi-robot by blockchain to combat COVID-19 and future pandemics. Some of the challenges of the proposed framework solutions are including (i) management blockchain lifecycles, (ii) ensuring that joining new robots into the blockchain network is to show the real quality of their data. Furthermore, the complexity is that it is challenging to evaluate robot sensing capabilities and data quality. The complexity may significantly impact robot resources and inaccurate data gathered by IoT devices.

### 2. OPPORTUNITIES

We observe some opportunities regarding combat COVID-19 in outdoor and hospital delivery systems such as smart homes and smart hospitals. Smart cities represent the key technology for combating COVID-19. As part of smart cities and smart hospitals, blockchain can be developed by deploying smart infrastructure of the environment (supported by smart IoT devices), where trust between multi-robot collaboration and smart infrastructure improves robot action in the same environment. Therefore, integration between smart IoT devices and multi-robot collaboration in a smart environment (smart hospital, smart city) can provide efficient data to multi-robot collaboration and validate the data stamps from multi-robot collaboration in the same environment.

## VIII. CONCLUSION

The combination of robots and blockchain plays a vital role in improving the healthcare domain in early diagnosis, quarantine time, and recovery of a disease outbreak. The number of infected cases of COVID-19 is reduced due to the increasing robot distributed functions for different purposes. Blockchain enables homogenous and heterogenous robots to combat COVID-19 collaboratively and efficiently by sharing information autonomously and accessing information of each other. Due to blockchain features enable multi-robot to log information (i.e., world states, time, location, resources, etc.) and deliver data and robot relation to all neighbors' robots. Therefore, blockchain empowers robots to operate safely, collaborate efficiently, act accurately and fast, and exchange behavior quickly. Multi-robot collaboration in blockchain network needs protocols (consensus algorithms) and techniques (sharding) to keep connectivity, avoid the collision, improve network partitioning, the response in real-time, and exchange behavior according to the environment. We discuss multi-robot collaboration in the blockchain network for controlling and improve decentralized to combat COVID-19. Applications of multi-robot collaboration for combating COVID-19 are discussed deeply in this paper with the proposed framework solution such as monitoring COVID-19, Outdoor E2E delivery system, and hospital E2E delivery system. Based on our proposed framework solutions, we provide potential research experimental future directions that can help develop blockchain-based multi-robot collaboration to combat COVID-19. Moreover, we motivate academic and industrial researchers to give more effort into the blockchain, multi-robot collaboration, and smart IoT technologies to combat COVID-19 and future pandemics.

## REFERENCES

[1] WHO, "Coronavirus disease (covid-19),", 2020, [online] available in:https://bit.ly/2ZU5x08, access on 15 Septemper, 2020.

[2] K. K.-W. To *et al.*, "Consistent detection of 2019 novel coronavirus in saliva," *Clinical Infectious Diseases,* 2020.

[3] E. Demaitre, "Coronavirus response growing from robotics companies,",2020, [online] availbale in: https://www.therobotreport.com/coronavirus-response-growing-robotics-companies/, access on 03 March, 2020.

[4] V. Chamola, V. Hassija, V. Gupta, and M. Guizani, "A Comprehensive Review of the COVID-19 Pandemic and the Role of IoT, Drones, AI, Blockchain, and 5G in Managing its Impact," *IEEE Access,* vol. 8, pp. 90225-90265, 2020.

[5] S. Alsamhi, O. Ma, and M. S. Ansari, "Convergence of Machine Learning and Robotics Communication in Collaborative Assembly: Mobility, Connectivity and Future Perspectives," *Journal of Intelligent & Robotic Systems,* pp. 1-26, 2019.

[6] S. H. Alsamhi, O. Ma, M. S. Ansari, and S. K. Gupta, "Collaboration of drone and internet of public safety things in smart cities: An overview of qos and network performance optimization," *Drones,* vol. 3, no. 1, p. 13, 2019.

[7] S. H. Alsamhi, O. Ma, and M. S. Ansari, "Survey on artificial intelligence based techniques for emerging robotic communication," *Telecommunication Systems,* vol. 72, no. 3, pp. 483-503, 2019.

[8] S. H. Alsamhi, O. Ma, M. S. Ansari, and F. A. Almalki, "Survey on collaborative smart drones and internet of things for improving smartness of smart cities," *Ieee Access,* vol. 7, pp. 128125-128152, 2019.

[9] A. Khawalid, D. Acristinii, H. van Toor, and E. C. Ferrer, "Grex: A decentralized hive mind," *Ledger,* vol. 4, 2019.

[10] R. Aragues, J. Cortes, and C. Sagues, "Distributed consensus on robot networks for dynamically merging feature-based maps," *IEEE Transactions on Robotics,* vol. 28, no. 4, pp. 840-854, 2012.

[11] G. P. Das, T. M. McGinnity, S. A. Coleman, and L. Behera, "A fast distributed auction and consensus process using parallel task allocation and execution," in *2011 IEEE/RSJ International Conference on Intelligent Robots and Systems,* 2011, pp. 4716-4721: IEEE.

[12] I. Navarro and F. Matía, "A framework for the collective movement of mobile robots based on distributed decisions," *Robotics and Autonomous Systems,* vol. 59, no. 10, pp. 685-697, 2011.

[13] S. Pourmehr, V. M. Monajjemi, R. Vaughan, and G. Mori, ""You two! Take off!": Creating, modifying and commanding groups of robots using face engagement and indirect speech in voice commands," in *2013 IEEE/RSJ International Conference on Intelligent Robots and Systems,* 2013, pp. 137-142: IEEE.

[14] E. C. Ferrer, "The blockchain: a new framework for robotic swarm systems," in *Proceedings of the future technologies conference,* 2018, pp. 1037-1058: Springer.

[15] V. K. Chattu, A. Nanda, S. K. Chattu, S. M. Kadri, and A. W. Knight, "The emerging role of blockchain technology applications in routine disease surveillance systems to strengthen global health

security," *Big Data and Cognitive Computing,* vol. 3, no. 2, p. 25, 2019.

[16] K. Salah, M. H. U. Rehman, N. Nizamuddin, and A. Al-Fuqaha, "Blockchain for AI: Review and open research challenges," *IEEE Access,* vol. 7, pp. 10127-10149, 2019.

[17] V. Strobel, E. Castelló Ferrer, and M. Dorigo, "Managing byzantine robots via blockchain technology in a swarm robotics collective decision making scenario," in *Proceedings of the 17th International Conference on Autonomous Agents and MultiAgent Systems,* 2018, pp. 541-549: International Foundation for Autonomous Agents and Multiagent Systems.

[18] M. G. M. M. Hasan, A. Datta, M. A. Rahman, and H. Shahriar, "Chained of things: A secure and dependable design of autonomous vehicle services," in *2018 IEEE 42nd Annual Computer Software and Applications Conference (COMPSAC),* 2018, vol. 2, pp. 498-503: IEEE.

[19] A. Kapitonov *et al.*, "Robotic Services for New Paradigm Smart Cities Based on Decentralized Technologies," *Ledger,* vol. 4, 2019.

[20] E. C. Ferrer, T. Hardjono, and A. Pentland, "Proceedings of the First Symposium on Blockchain and Robotics, MIT Media Lab, 5 December 2018," *Ledger,* vol. 4, 2019.

[21] A. Kapitonov, S. Lonshakov, A. Krupenkin, and I. Berman, "Blockchain-based protocol of autonomous business activity for multi-agent systems consisting of UAVs," in *2017 Workshop on Research, Education and Development of Unmanned Aerial Systems (RED-UAS),* 2017, pp. 84-89: IEEE.

[22] D. S. W. Ting, L. Carin, V. Dzau, and T. Y. Wong, "Digital technology and COVID-19," *Nature medicine,* vol. 26, no. 4, pp. 459-461, 2020.

[23] D. Nguyen, M. Ding, P. N. Pathirana, and A. Seneviratne, "Blockchain and AI-based Solutions to Combat Coronavirus (COVID-19)-like Epidemics: A Survey," *10.36227/techrxiv.12121962.v1,* 2020.

[24] S. Agarwal, N. S. Punn, S. K. Sonbhadra, P. Nagabhushan, K. K. Pandian, and P. Saxena, "Unleashing the power of disruptive and emerging technologies amid COVID 2019: A detailed review," *arXiv preprint arXiv:2005.11507,* 2020.

[25] N. Saeed, A. Bader, T. Y. Al-Naffouri, and M.-S. Alouini, "When Wireless Communication Faces COVID-19: Combating the Pandemic and Saving the Economy," *arXiv preprint arXiv:2005.06637,* 2020.

[26] N. Melluso *et al.*, "Lights and shadows of COVID-19, Technology and Industry 4.0," *arXiv preprint arXiv:2004.13457,* 2020.

[27] M. Nasajpour, S. Pouriyeh, R. M. Parizi, M. Dorodchi, M. Valero, and H. R. Arabnia, "Internet of Things for Current COVID-19 and Future Pandemics: An Exploratory Study," *arXiv preprint arXiv:2007.11147,* 2020.

[28] V. Strobel, E. Castelló Ferrer, and M. Dorigo, "Blockchain technology secures robot swarms: A comparison of consensus protocols and their resilience to Byzantine robots," *Frontiers in Robotics and AI,* vol. 7, p. 54, 2020.

[29] T. Alam, "Internet of Things and Blockchain-based framework for Coronavirus (Covid-19) Disease," *Available at SSRN 3660503,* 2020.

[30] M. Tavakoli, J. Carriere, and A. Torabi, "Robotics, smart wearable technologies, and autonomous intelligent systems for healthcare during the COVID-19 pandemic: An analysis of the state of the art and future vision," *Advanced Intelligent Systems,* p. 2000071, 2020.

[31] Z. H. Khan, A. Siddique, and C. W. Lee, "Robotics Utilization for Healthcare Digitization in Global COVID-19 Management," *International Journal of Environmental Research and Public Health,* vol. 17, no. 11, p. 3819, 2020.

[32] G. Fragkos, E. E. Tsiropoulou, and S. Papavassiliou, "Disaster management and information transmission decision-making in public safety systems," in *2019 IEEE Global Communications Conference (GLOBECOM),* 2019, pp. 1-6: IEEE.

[33] X.-L. Huang, X. Ma, and F. Hu, "Machine learning and intelligent communications," *Mobile Networks and Applications,* vol. 23, no. 1, pp. 68-70, 2018.

[34] C. Singhal and S. De, *Resource allocation in next-generation broadband wireless access networks*. IGI Global, 2017.

[35] P. Vamvakas, E. E. Tsiropoulou, M. Vomvas, and S. Papavassiliou, "Adaptive power management in wireless powered communication networks: A user-centric approach," in *2017 IEEE 38th Sarnoff Symposium,* 2017, pp. 1-6: IEEE.

[36] S. Contreras, H. A. Villavicencio, D. Medina-Ortiz, J. P. Biron-Lattes, and Á. Olivera-Nappa, "A multi-group SEIRA model for the spread of COVID-19 among heterogeneous populations," *Chaos, Solitons & Fractals,* p. 109925, 2020.

[37] N. Crokidakis, "COVID-19 spreading in Rio de Janeiro, Brazil: do the policies of social isolation really work?," *Chaos, Solitons & Fractals,* p. 109930, 2020.

[38] M. S. Abdo, K. Shah, H. A. Wahash, and S. K. Panchal, "On a comprehensive model of the novel coronavirus (COVID-19) under Mittag-Leffler derivative," *Chaos, Solitons & Fractals,* p. 109867, 2020.

[39] S. Boccaletti, W. Ditto, G. Mindlin, and A. Atangana, "Modeling and forecasting of epidemic spreading: The case of Covid-19 and beyond," *Chaos, Solitons, and Fractals,* vol. 135, p. 109794, 2020.

[40] Q. Xiaoxia, "How next-generation information technologies tackled COVID-19 in China," in "weforum.org/agenda/2020/04/how-next-generation-information-technologies-tackled-covid-19-in-china/," 2020.

[41] A. Kalla, T. Hewa, R. Mishra, M. Ylianttila, and M. Liyanage, "The Role of Blockchain to Fight Against COVID-19," *IEEE Engineering Management Review,* vol. PP, 08/01 2020.

[42] V. Lopes and L. A. Alexandre, "An overview of blockchain integration with robotics and artificial intelligence," *arXiv preprint arXiv:1810.00329,* 2018.

[43] R. Gupta, A. Kumari, and S. Tanwar, "A taxonomy of blockchain envisioned edge-as-a-connected autonomous vehicles," *Transactions on Emerging Telecommunications Technologies,* p. e4009.

[44] E. Otiede *et al.*, "Delivering Medical Products to Quarantined Regions Using Unmanned Aerial Vehicles," *Journal of Applied Mechanics Engineering,* vol. 6, no. 1, p. 244, 2017.

[45] Y. Zheng, Y. Zhu, and L. Wang, "Consensus of heterogeneous multi-agent systems," *IET Control Theory & Applications,* vol. 5, no. 16, pp. 1881-1888, 2011.

[46] J. P. Queralta and T. Westerlund, "Managing Collaboration in Heterogeneous Swarms of Robots with Blockchains," *arXiv preprint arXiv:1912.01711,* 2019.

[47] G. Valentini, D. Brambilla, H. Hamann, and M. Dorigo, "Collective perception of environmental features in a robot swarm," in *International Conference on Swarm Intelligence,* 2016, pp. 65-76: Springer.

[48] I. Afanasyev *et al.*, "Towards blockchain-based multi-agent robotic systems: Analysis, classification and applications," *arXiv preprint arXiv:1907.07433,* 2019.

[49] M. U. Javed, M. Rehman, N. Javaid, A. Aldegheishem, N. Alrajeh, and M. Tahir, "Blockchain-Based Secure Data Storage for Distributed Vehicular Networks," *Applied Sciences,* vol. 10, no. 6, p. 2011, 2020.

[50] A. A. Monrat, O. Schelén, and K. Andersson, "A survey of blockchain from the perspectives of applications, challenges, and opportunities," *IEEE Access,* vol. 7, pp. 117134-117151, 2019.

[51] Z. Zheng, S. Xie, H.-N. Dai, X. Chen, and H. Wang, "Blockchain challenges and opportunities: A survey," *International Journal of Web and Grid Services,* vol. 14, no. 4, pp. 352-375, 2018.

[52] M. Dorigo, "Blockchain Technology for Robot Swarms: A Shared Knowledge and Reputation Management System for Collective Estimation," in *Swarm Intelligence: 11th International Conference, ANTS 2018, Rome, Italy, October 29–31, 2018, Proceedings*, 2018, vol. 11172, p. 425: Springer.

[53] I. Zikratov, O. Maslennikov, I. Lebedev, A. Ometov, and S. Andreev, "Dynamic trust management framework for robotic multi-agent systems," in *Internet of Things, Smart Spaces, and Next Generation Networks and Systems*: Springer, 2016, pp. 339-348.

[54] E. C. Ferrer, O. Rudovic, T. Hardjono, and A. Pentland, "Robochain: A secure data-sharing framework for human-robot interaction," *arXiv preprint arXiv:1802.04480,* 2018.

[55] R. Skowroński, "The open blockchain-aided multi-agent symbiotic cyber–physical systems," *Future Generation Computer Systems,* vol. 94, pp. 430-443, 2019.

[56] J. P. Queralta and T. Westerlund, "Blockchain-powered collaboration in heterogeneous swarms of robots," *Frontiers in Robotics and AI,* 2020.

[57] W. Cai, W. Jiang, K. Xie, Y. Zhu, Y. Liu, and T. Shen, "Dynamic reputation–based consensus mechanism: Real-time transactions for energy blockchain," *International Journal of Distributed Sensor Networks,* vol. 16, no. 3, p. 1550147720907335, 2020.

[58] J. A. Tran, G. S. Ramachandran, P. M. Shah, C. B. Danilov, R. A. Santiago, and B. Krishnamachari, "SwarmDAG: A Partition Tolerant Distributed Ledger Protocol for Swarm Robotics," *Ledger,* vol. 4, 2019.

[59] T. McConaghy *et al.*, "Bigchaindb: a scalable blockchain database," *white paper, BigChainDB,* 2016.

[60] A. Kiayias and G. Panagiotakos, "Speed-Security Tradeoffs in Blockchain Protocols," *IACR Cryptology ePrint Archive,* vol. 2015, p. 1019, 2015.

[61] M. Fernandes and L. A. Alexandre, "Robotchain: Using tezos technology for robot event management," *Ledger,* vol. 4, 2019.

[62] A. K. R. Venkatapathy and M. ten Hompel, "A Decentralized Context Broker Using Byzantine Fault Tolerant Consensus," *Ledger,* vol. 4, 2019.

[63] S. Falcone, J. Zhang, A. Cameron, and A. Abdel-Rahman, "Blockchain Design for an Embedded System," *Ledger,* vol. 4, 2019.

[64] Dezeen, "Drones and self-driving robots used to fight coronavirus in China,"2020, [online] available in: https://www.dezeen.com/2020/02/20/drones-robots-coronavirus-china-technology/, access on 21 February, 2020.

[65] J. Happich, "5G edge patrol robots deployed in China to detect Covid-19 cases," 2020, [online] available in: https://www.eenewseurope.com/news/5g-edge-patrol-robots-deployed-china-detect-covid-19-cases/page/0/1, accessed on 29 March, 2020.

[66] Z. Doffman, "This New Coronavirus Spy Drone Will Make Sure You Stay Home," 2020, [online], available in: https://www.forbes.com/sites/zakdoffman/2020/03/05/meet-the-coronavirus-spy-drones-that-make-sure-you-stay-home/#43a7c9e51669, access on 23 March, 2020.

[67] W. Robotics, "Drones and the Coronavirus: Do These Applications Make Sense,",2020, [online] available in: https://blog.werobotics.org/2020/04/09/drones-coronavirus-no-sense/, access on 12 Septemper, 2020.

[68] P. Singh, "Thermal scan: Drones check people for fever in Delhi," ,2020, [online] available in: https://timesofindia.indiatimes.com/city/delhi/thermal-scan-drones-check-people-for-fever/articleshow/75088774.cms, access on 13 April, 2020.

[69] M. Griffin, "Thermal imaging enabled drones help Chinese identify people with Covid-19,", 2020, [online] available in: https://www.fanaticalfuturist.com/2020/03/thermal-imaging-enabled-drones-help-chinese-identify-people-with-covid-19/, access on 13 March, 2020.

[70] E. D. Otiede *et al.*, "Delivering Medical Products to Quarantined Regions Using Unmanned Aerial Vehicles," *Journal of Applied Mechanics Engineering,* vol. 6, no. 1, p. 244, 2017.

[71] A. Hardy, M. Makame, D. Cross, S. Majambere, and M. Msellem, "Using low-cost drones to map malaria vector habitats," *Parasites & vectors,* vol. 10, no. 1, p. 29, 2017.

[72] A. Sathyan and O. Ma, "Collaborative Control of Multiple Robots Using Genetic Fuzzy Systems," *Robotica,* vol. 37, no. 11, pp. 1922-1936, 2019.

[73] Q. Li, R. Gravina, Y. Li, S. H. Alsamhi, F. Sun, and G. Fortino, "Multi-user Activity Recognition: Challenges and Opportunities," *Information Fusion,* 2020.

[74] E. Ackerman, "Autonomous Robots Are Helping Kill Coronavirus in Hospitals," *IEEE Spectrum,* 2020.

[75] M. Editors. (2020). *Automated robot takes swabs for safe covid-19 testing.*

[76] M. Tao. (2020). *Cobots v covid: How universal robots and others are helping in the fight against coronavirus.*

[77] D. Calvaresi, A. Dubovitskaya, J. P. Calbimonte, K. Taveter, and M. Schumacher, "Multi-agent systems and blockchain: Results from a systematic literature review," in *International Conference on Practical Applications of Agents and Multi-Agent Systems,* 2018, pp. 110-126: Springer.

[78] I. Afanasyev, A. Kolotov, R. Rezin, K. Danilov, A. Kashevnik, and V. Jotsov, "Blockchain solutions for multi-agent robotic systems: Related work and open questions," in *Proceedings of the 24th Conference of Open Innovations Association FRUCT,* 2019, p. 76: FRUCT Oy.

[79] J. Peña Queralta and T. Westerlund, "Blockchain-Powered Collaboration in Heterogeneous Swarms of Robots," *arXiv,* p. arXiv: 1912.01711, 2019.

[80] G. Valentini, E. Ferrante, and M. Dorigo, "The best-of-n problem in robot swarms: Formalization, state of the art, and novel perspectives," *Frontiers in Robotics and AI,* vol. 4, p. 9, 2017.

[81] M. Andoni *et al.*, "Blockchain technology in the energy sector: A systematic review of challenges and opportunities," *Renewable and Sustainable Energy Reviews,* vol. 100, pp. 143-174, 2019.

[82] X. Chuanjiao, "Robots and drones join the battle against virus,",2020, [online] available in: https://www.chinadaily.com.cn/a/202002/14/WS5e46af05a3101282 17277b28.html, access on 29 Febratury, 2020.

[83] A. Hand, "COVID-19 Provides Use Cases for Mobile Robotics,", 2020, [online] available in: https://www.healthcarepackaging.com/covid-19/article/21126536/covid19-provides-use-cases-for-mobile-robotics, access on 02 April, 2020.

[84] A. Naidoo, "The Heat: Containing coronavirus – China and world's response,",2020, [online] available in: https://www.chinadaily.com.cn/a/202002/14/WS5e46af05a3101282 17277b28.html, access on 12 June 2020.

[85] B. AMES, "Robot manufacturers join Covid-19 fight with disinfectant-bots,",2020, [online] available in: https://www.dcvelocity.com/articles/45599-robot-manufacturers-join-covid-19-fight-with-disinfectant-bots,"access on 4 April, 2020.

[86] L. M. Goodman, "Tezos: A self-amending crypto-ledger position paper," *Aug,* vol. 3, p. 2014, 2014.

[87] N. Szabo, "Micropayments and mental transaction costs," in *2nd Berlin Internet Economics Workshop,* 1999.

[88] D. Yaga, P. Mell, N. Roby, and K. Scarfone, "Blockchain technology overview," *arXiv preprint arXiv:1906.11078,* 2019.

[89] B. Hill, S. Chopra, P. Valencourt, and N. Prusty, *Blockchain Developer's Guide: Develop smart applications with Blockchain technologies-Ethereum, JavaScript, Hyperledger Fabric, and Corda.* Packt Publishing Ltd, 2018.

[90] C. Dannen, *Introducing Ethereum and solidity.* Springer, 2017.

[91] J. Qin, Q. Ma, Y. Shi, and L. Wang, "Recent advances in consensus of multi-agent systems: A brief survey," *IEEE Transactions on Industrial Electronics,* vol. 64, no. 6, pp. 4972-4983, 2016.

[92] W. Wang *et al.*, "A survey on consensus mechanisms and mining strategy management in blockchain networks," *IEEE Access,* vol. 7, pp. 22328-22370, 2019.

[93] D. Vujičić, D. Jagodić, and S. Ranđić, "Blockchain technology, bitcoin, and Ethereum: A brief overview," in *2018 17th international symposium infoteh-jahorina (infoteh),* 2018, pp. 1-6: IEEE.

[94] M. Vukolić, "The quest for scalable blockchain fabric: Proof-of-work vs. BFT replication," in *International workshop on open problems in network security,* 2015, pp. 112-125: Springer.

[95] L. Luu, V. Narayanan, C. Zheng, K. Baweja, S. Gilbert, and P. Saxena, "A secure sharding protocol for open blockchains," in *Proceedings of the 2016 ACM SIGSAC Conference on Computer and Communications Security,* 2016, pp. 17-30.

[96] D. s. den. (2016). *Using Merklix tree to shard block validation.*

[97] B. Qin, J. Huang, Q. Wang, X. Luo, B. Liang, and W. Shi, "Cecoin: A decentralized PKI mitigating MitM attacks," *Future Generation Computer Systems,* vol. 107, pp. 805-815, 2020.

[98] V. Buterin, "A next-generation smart contract and decentralized application platform," *white paper,* vol. 3, no. 37, 2014.

[99] S. E. F. e. al. (2018). *Ethereum 2.0 Specifications*.

[100] P. S, "The tangle," 2018, [online] available in: https://www.iota.org/research/academic-papers, access on 25 April, 2020.

[101] L. Baird, "The swirlds hashgraph consensus algorithm: Fair, fast, byzantine fault tolerance," *Swirlds, Inc. Technical Report SWIRLDS-TR-2016,* vol. 1, 2016.